\documentclass{article}

\usepackage{microtype}
\usepackage{graphicx}
\usepackage{subcaption}
\usepackage{booktabs} 
\usepackage[table,xcdraw]{xcolor}

\usepackage{hyperref}

\usepackage[accepted]{icml2024}

\usepackage{amsmath}
\usepackage{amssymb}
\usepackage{mathtools}
\usepackage{amsthm}

\usepackage[capitalize,noabbrev]{cleveref}

\theoremstyle{plain}

\theoremstyle{definition}

\theoremstyle{remark}

\usepackage[textsize=tiny]{todonotes}

\usepackage{enumitem}

\usepackage{amsmath, amsfonts, bm}

\newcommand{\mlp}{\text{MLP}}

\def\eqref#1{equation~\ref{#1}}

\def\1{\bm{1}}

\def\vzero{{\bm{0}}}
\def\vone{{\bm{1}}}

\def\vtheta{{\bm{\theta}}}

\def\vh{{\bm{h}}}

\def\vm{{\bm{m}}}

\def\vp{{\bm{p}}}

\def\vu{{\bm{u}}}

\def\vx{{\bm{x}}}

\def\vz{{\bm{z}}}

\def\mA{{\bm{A}}}

\def\mI{{\bm{I}}}

\def\mM{{\bm{M}}}

\def\mR{{\bm{R}}}
\def\mS{{\bm{S}}}
\def\mT{{\bm{T}}}
\def\mU{{\bm{U}}}

\def\mX{{\bm{X}}}

\def\mZ{{\bm{Z}}}

\DeclareMathAlphabet{\mathsfit}{\encodingdefault}{\sfdefault}{m}{sl}
\SetMathAlphabet{\mathsfit}{bold}{\encodingdefault}{\sfdefault}{bx}{n}

\def\gB{{\mathcal{B}}}

\def\gG{{\mathcal{G}}}

\def\gL{{\mathcal{L}}}

\def\gN{{\mathcal{N}}}
\def\gO{{\mathcal{O}}}

\def\gU{{\mathcal{U}}}
\def\gV{{\mathcal{V}}}

\def\sR{{\mathbb{R}}}

\newcommand{\tuple}[1]{{\left\langle #1 \right\rangle}}

\DeclareMathOperator*{\softmax}{\mathrm{softmax}}

\usepackage[acronym, nohypertypes={acronym}]{glossaries}
\glsdisablehyper

\newacronym{gnn}{GNN}{graph neural network}
\newacronym{stgnn}{STGNN}{spatiotemporal graph neural network}
\newacronym{mlp}{MLP}{multilayer perceptron}
\newacronym{cnn}{CNN}{convolutional neural network}
\newacronym{tcn}{TCN}{temporal convolutional network}
\newacronym{rnn}{RNN}{recurrent neural network}
\newacronym{gru}{GRU}{Gated Recurrent Unit}

\newacronym{mp}{MP}{message passing}
\newacronym{tmp}{TMP}{temporal message-passing}
\newacronym{smp}{SMP}{spatial message-passing}
\newacronym{stmp}{STMP}{spatiotemporal message-passing}

\newacronym{mae}{MAE}{mean absolute error}

\newacronym{sn}{SN}{sensor network}
\newacronym{iot}{IoT}{Internet of Things}
\newacronym{mso}{GraphMSO}{Graph Multiple Superimposed Oscillators}

\newacronym{tts}{TTS}{\emph{time-then-space}}
\newacronym{tas}{T\&S}{\emph{time-and-space}}

\newacronym{mcar}{MCAR}{missing completely at random}
\newacronym{mar}{MAR}{missing at random}
\newacronym{mnar}{MNAR}{missing not at random}

\newglossaryentry{tmpd}{name=\textsc{TMP-D},description=}
\newglossaryentry{smpd}{name=\textsc{SMP-D},description=}

\newacronym{model}{HD-TTS}{Hierarchical Downsampling Time-Then-Space}

\newglossaryentry{ttsimp}{name=TTS-IMP,description=}
\newglossaryentry{ttsamp}{name=TTS-AMP,description=}
\newglossaryentry{tasimp}{name=T\&S-IMP,description=}
\newglossaryentry{tasamp}{name=T\&S-AMP,description=}

\newglossaryentry{fcrnn}{name=FC-RNN,description=}
\newglossaryentry{fcgru}{name=FCGRU,description=}
\newglossaryentry{grud}{name=GRU-D,description=}
\newglossaryentry{fcgrud}{name=FCGRU-D,description=}
\newglossaryentry{grui}{name=GRU-I,description=}
\newglossaryentry{grin}{name=GRIN-P,description=}
\newglossaryentry{dcrnn}{name=DCRNN,description=}
\newglossaryentry{agcrn}{name=AGCRN,description=}
\newacronym{gwnet}{GWNet}{Graph WaveNet}

\newglossaryentry{la}{name=METR-LA,description=}
\newglossaryentry{bay}{name=PEMS-BAY,description=}
\newglossaryentry{air}{name=AQI,description=}
\newglossaryentry{pvus}{name=PV-US,description=}
\newglossaryentry{engrad}{name=EngRAD,description=}
\newglossaryentry{msods}{name=GraphMSO,description=}

\newglossaryentry{point}{name=Point,description=}
\newglossaryentry{block}{name=Block--T,description=}
\newglossaryentry{block_prop}{name=Block--ST,description=}

\newcommand{\appref}[1]{\hyperref[#1]{Appendix~\ref*{#1}}}

\newcommand{\autorefp}[1]{(\autoref{#1})}
\newcommand{\autorefseq}[2]{\autoref{#1}--\ref{#2}}
\newcommand{\autorefseqp}[2]{(\autoref{#1}--\ref{#2})}

\newcommand{\geqzero}{\scalebox{.7}{$\geq 0$}}
\DeclareMathSymbol{\shortminus}{\mathbin}{AMSa}{"39}

\begin{document}

\twocolumn[
\icmltitle{Graph-based Forecasting with Missing Data through~Spatiotemporal~Downsampling}



\icmlsetsymbol{equal}{*}

\begin{icmlauthorlist}
\icmlauthor{Ivan Marisca}{usi}
\icmlauthor{Cesare Alippi}{usi,poli}
\icmlauthor{Filippo Maria Bianchi}{uit,norce}
\end{icmlauthorlist}

\icmlaffiliation{usi}{IDSIA USI-SUPSI, Universit\`a della Svizzera italiana}
\icmlaffiliation{poli}{Politecnico di Milano}
\icmlaffiliation{uit}{Dept. of Mathematics and Statistics, UiT the Arctic University of Norway}
\icmlaffiliation{norce}{NORCE, Norwegian Research Centre AS}

\icmlcorrespondingauthor{Ivan Marisca}{ivan.marisca@usi.ch}
\icmlcorrespondingauthor{Filippo Maria Bianchi}{filippo.m.bianchi@uit.no}

\icmlkeywords{Time series, Forecasting, Recurrent Neural Networks, Graph Neural Networks, Spatiotemporal, Graph Pooling.}

\vskip 0.3in
]



\printAffiliationsAndNotice{}  


\begin{abstract}
    Given a set of synchronous time series, each associated with a sensor-point in space and characterized by inter-series relationships, the problem of spatiotemporal forecasting consists of predicting future observations for each point. 
\Acrlongpl{stgnn} achieve striking results by representing the relationships across time series as a graph. Nonetheless, most existing methods rely on the often unrealistic assumption that inputs are always available and fail to capture hidden spatiotemporal dynamics when part of the data is missing. In this work, we tackle this problem through hierarchical spatiotemporal downsampling. The input time series are progressively coarsened over time and space, obtaining a pool of representations that capture heterogeneous temporal and spatial dynamics. Conditioned on observations and missing data patterns, such representations are combined by an interpretable attention mechanism to generate the forecasts. Our approach outperforms state-of-the-art methods on synthetic and real-world benchmarks under different missing data distributions, particularly in the presence of contiguous blocks of missing values.
\end{abstract}


\section{Introduction}
\label{sec:intro}

Time-series analysis and forecasting often deal with high-dimensional data acquired by \glspl{sn}, a broad term for systems that collect (multivariate) measurements over time at different spatial locations. Examples include systems monitoring air quality, where each sensor records air pollutants' concentrations, or traffic, where sensors track vehicles' flow or speed. Usually, data are sampled regularly over time and synchronously across the sensors, which are often characterized by strong correlations and dependencies between each other, i.e., across the \textit{spatial} dimension. For this reason, a prominent deep learning approach is to consider the time series and their relationships as graphs and to process them with architectures that combine \glspl{gnn}~\cite{battaglia2018relational, bronstein2021geometric} with sequence-processing operators~\cite{hochreiter1997long, borovykh2017conditional}. These architectures are known as \glspl{stgnn}~\cite{jin2023survey}.

\begin{figure}[t]
    \begin{center}
    \includegraphics[width=\linewidth]{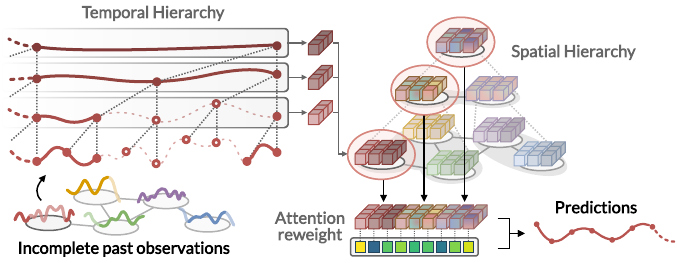}
    \caption{Overview of the proposed framework. The hierarchical design allows us to learn a pool of multi-scale spatiotemporal representations. Conditioned on the data and the missing value pattern, the attention mechanism dynamically combines representation from different scales to compute the predictions.}
    \label{fig:overview}
    \end{center}
\end{figure}

A notable limit of most existing \glspl{stgnn} is the assumption that inputs are complete and regular sequences.  However, real-world \glspl{sn} are prone to failures and faults, resulting eventually in missing values in the collected time series.  
When missing data occurs randomly and sporadically, the localized processing imposed by the inductive biases in \glspl{stgnn} acts as an effective regularization, exploiting observations close in time and space to the missing one~\cite{cini2022filling}.
Challenges arise when data are missing in larger and contiguous blocks, with gaps that occur in consecutive time steps and are spatially proximate. In \glspl{sn}, this might be due to a sensor failure lasting for multiple time lags or problems affecting a whole portion of the network. 
In such scenarios, reaching valid observations that may be significantly distant in time and space, yet relevant for capturing the underlying dynamics, would require additional layers of processing~\cite{marisca2022learning}. Nonetheless, deep processing may attenuate faster dynamics, impairing the network's ability to rely on local information, if present~\cite{rusch2023survey}.
The necessity to expand the network's receptive field should be, therefore, contingent on the availability of input data. As such, it is crucial to tailor the processing strategy according to the dynamics in the input data and the specific patterns of missing information.

In this work, we propose a deep learning framework for graph-based forecasting of time series with missing data that computes representations at multiple spatiotemporal scales and weighs them conditioned on the observations and missing data pattern at hand.
To this end, we rely on operators that progressively reduce the data granularity across both temporal and spatial dimensions (see~\autoref{fig:overview}). 
In time, we interleave downsampling within temporal processing layers, handling varying noise levels and isolating specific temporal dynamics. 
In space, we use graph pooling~\cite{grattarola2024understanding} to obtain a hierarchy of coarsened graphs that gradually distill the global information necessary for compensating localized gaps in the data.
The proposed framework adopts a \emph{time-then-space}~\cite{gao2022equivalence} hierarchical design, which efficiently handles representations at multiple scales by increasing the receptive field while limiting the number of parameters and the amount of computation. 
The hierarchical representations learned by our model are then combined by a soft attention mechanism~\cite{vaswani2017attention}, whose scores offer a natural interpretability tool to inspect the model's behavior in function of the data.

We compare our approach against state-of-the-art methods in both synthetic and real-world benchmark datasets, showing remarkable improvements in efficiency and performance, particularly with large blocks of missing values. Notably, we introduce two new datasets and an experimental setting specifically designed to reflect typical missing data patterns in the spatiotemporal domain. This contribution addresses a substantial gap in the existing literature by providing a controlled environment for testing the performance and understanding the behavior of complex spatiotemporal models.

\section{Preliminaries and Problem Formulation}
\label{sec:preliminaries}

We represent each sensor in a \gls{sn} (i.e., a point in space) as a node in the set $\gV$, with $|\gV| = N$. 
We model dyadic and possibly asymmetric relationships between sensors with a weighted adjacency matrix $\mA \in \sR_{\geqzero}^{N \times N}$, where each non-zero entry $a^{ij}$ is the nonnegative weight of the directed edge from the $i$-th to the $j$-th node.
We denote by $\vx^i_t \in \sR^{d_x}$, the $d_x$-dimensional observation collected by the $i$-th node at time step $t$, with $\mX_t \in \sR^{N\times d_x}$ representing all the observations collected synchronously in the \gls{sn}. We use the notation $\mX_{t:t+T}$ to indicate the sequence of $T$ observations in the time interval $[t, t+T)$.
We represent node-level exogenous variable with matrix $\mU_t \in \sR^{N\times d_u}$ (e.g., date/time information, external events). To model the presence of missing data, we associate every node observation with a binary mask $\vm^i_t \in \{0, 1\}^{d_x}$, whose elements are nonzero whenever the corresponding channel in $\vx^i_t$ is valid. Notably, we do not make any assumption on the missing data distribution and we consider $\vu^i_t$ always observed, regardless of $\vm^i_t$. Finally, we use the tuple $\gG_t=\tuple{\mX_t, \mM_t, \mU_t, \mA}$ to denote all information available at time step $t$.

Given a \textit{window} $\gG_{t-W:t}$ of $W$ past observations, the problem of \emph{spatiotemporal forecasting} consists of predicting an \textit{horizon} of $H$ future observations for each node $i \in \gV$:
\begin{equation}
    \hat{\vx}^i_{t:t+H} = f(\gG_{t-W:t}) .
\end{equation}
As observations might also be missing in the ground-truth data, to measure forecasting accuracy we average an element-wise loss function $\ell$ (e.g., absolute or squared error) over only valid values, i.e.,
\begin{gather}
    \gL_{t:t+H} = \sum_{h=t}^{t+H-1}\sum_{i=1}^{N}\frac{\left\| \vm^i_h \odot \ell\left(\hat \vx_h^i, \vx_h^i\right) \right\|_1}{\left\| \vm^i_h\right\|_1} \label{eq:loss},
\end{gather}
where $\odot$ is the Hadamard product.

\subsection{Spatiotemporal Message Passing}
\label{sec:stmp}

The cornerstone operator of an \gls{stgnn} is the \gls{stmp} layer~\cite{cini2023taming}, which computes nodes' features at the $l$-th layer as:
\begin{equation}
\label{eq:stmp}
    \vx^{i,l}_{t} = \gamma^{l} \left( \vx^{i,l \shortminus 1}_{\leq t}, \operatorname*{\textsc{Aggr}}_{j \in \mathcal{N}(i)} \phi^{l} \left( \vx^{i,l \shortminus 1}_{\leq t}, \vx^{j,l \shortminus 1}_{\leq t}, a^{ji} \right) \right)
\end{equation}
where \textsc{Aggr} is a differentiable, permutation invariant \textit{aggregation} function, e.g., sum or mean, and $\gamma^{l}$ and $\phi^{l}$ are differentiable \textit{update} and \textit{message} functions, respectively. 
Whenever $\gamma^{l}$ and $\phi^{l}$ are such that temporal and spatial processing cannot be factorized in two distinguished operations, the \gls{stgnn} is said to follow a \gls{tas} paradigm~\cite{gao2022equivalence, cini2023graph}. 
In the \gls{tts} approach, instead, the input sequences are first encoded in a vector by \gls{tmp} layers before being propagated on the graph by \gls{smp} layers. 

We use the term \gls{tmp} to refer broadly to any deep learning operator enabling the exchange of information along the temporal dimension. Most \gls{tmp} operators can be categorized as recurrent or convolutional. 
\Glspl{rnn}~\cite{elman1990finding} process sequential data of varying lengths in a recursive fashion, by maintaining a memory of previous inputs:
\begin{equation}
    \label{eq:tmp_rnn}
    \vx^{i,l}_{t} = \gamma^{l} \left( \vx^{i,l \shortminus 1}_{t}, \phi^{l} \left( \vx^{i,l \shortminus 1}_{t}, \vx^{i,l}_{t-1} \right) \right). 
\end{equation}
In modern \glspl{rnn}, gating mechanisms are used to cope with vanishing (or exploding) gradients that hinder learning long-range dependencies~\cite{hochreiter1997long, cho2014learning}.
Convolutional \gls{tmp} operators, instead, learn causal filters conditioned on a sequence of previous observations:
\begin{equation}
     \label{eq:tmp_tcn}
    \vx^{i,l}_{t} = \gamma^{l} \left( \vx^{i,l \shortminus 1}_{t}, \operatorname*{\textsc{Aggr}}_{k > 0} \phi^{l} \left( \vx^{i,l \shortminus 1}_{t}, \vx^{i,l \shortminus 1}_{t-k} \right) \right) .
\end{equation}
\Glspl{tcn}~\cite {borovykh2017conditional, oord2016wavenet} and attention-based methods~\cite{zhou2021informer} follow this approach. An advantage of convolutional \gls{tmp} is its ability to be executed in parallel along the temporal axis~\autorefp{eq:tmp_tcn}, offering computational benefits over recurrent \gls{tmp}~\autorefp{eq:tmp_rnn}.

Instead, the \gls{smp} operator~\cite{gilmer2017neural} can be described as
\begin{equation}
    \label{eq:smp}
    \vx^{i,l}_{t} = \gamma^{l} \left( \vx^{i,l \shortminus 1}_{t}, \operatorname*{\textsc{Aggr}}_{j \in \mathcal{N}(i)} \phi^{l} \left( \vx^{i,l \shortminus 1}_{t}, \vx^{j,l \shortminus 1}_{t}, a^{ji} \right) \right) . 
\end{equation}
If the messages depend on the receiver node's features $\vx^{i,l \shortminus 1}_{t}$ the \gls{smp} operator is called \textit{anisotropic}~\cite{dwivedi2020benchmarking}. 
Conversely, if the message function depends only on the source node's features $\vx^{j,l \shortminus 1}_{t}$ and the edge weight $a^{ji}$, the \gls{smp} operator is said to be \textit{isotropic} or \textit{convolutional}~\cite{bronstein2021geometric}. 

Notably, the \gls{tmp} operators~\autorefseqp{eq:tmp_rnn}{eq:tmp_tcn} and the \gls{smp} operator~\autorefp{eq:smp} are specific instances of the \gls{stmp} operator~\autorefp{eq:stmp}, underlining the factorization of operations within \gls{tts} models compared to the \gls{tas} approach.
Since \gls{tts} models perform \gls{smp}~--~an onerous operation~--~on a single graph rather than a sequence of graphs, they are more efficient than \gls{tas} models. Nonetheless, their uncoupled temporal and spatial processing reduces the flexibility in how information is propagated, compared to the \gls{tas} processing~\cite{cini2023graph}. In the latter, indeed, it is possible to gradually account for more information while processing the temporal dimension and allow the receptive field to grow with the sequence length.

\subsection{Spatiotemporal Downsampling}
\label{sec:downsamplig}

Downsampling in temporal data is a common operation, often used to reduce the sample complexity or filter out noisy measurements~\cite{harris2022multirate}. In classical signal processing, it is implemented by applying a low-pass filter and then keeping only 1-every-$k$ samples, with $k$ being the \textit{downsampling factor}~\cite{strang1996wavelets}. 
This approach is replicated by strided operations in \gls{tmp}~\cite{yu2016multi, oord2016wavenet, chang2017dilated}, which exploit the structural regularity of temporal data. 
More generally, downsampling a sequence from $W_{l \shortminus 1}$ to  $W_{l}$ time steps can be conveniently expressed by a \textit{temporal downsampling matrix} $\mT_l \in \sR^{W_{l} \times W_{l \shortminus 1}}$. 
For example, $\mT_l = [\mI_{W_{l \shortminus 1}}]_{::k}$, i.e., an identity matrix without the rows associated with the decimated time steps, can be applied to keep the samples associated with every $k$-th time step.

Being non-Euclidean structures, the concept of downsampling for graphs is less straightforward and tied to the procedure for graph coarsening. In the \glspl{gnn} literature, the latter is known as \emph{graph pooling}.
Given a graph $\mA^{\tuple{k \shortminus 1}}\in\sR_{\geqzero}^{N_{k \shortminus 1} \times N_{k \shortminus 1}}$ with features $\mX^{\tuple{k \shortminus 1}} \in \sR^{N_{k \shortminus 1} \times d_x}$ on the nodes, the Select-Reduce-Connect (SRC) framework by~\citet{grattarola2024understanding} expresses a graph pooling operator $\texttt{POOL}: (\mA^{\tuple{k \shortminus 1}},\mX^{\tuple{k \shortminus 1}}) \mapsto (\mA^{\tuple{k}}, \mX^{\tuple{k}})$ as the combination of three functions:
\begin{itemize}[leftmargin=1.5em, itemsep=.1em, topsep=0pt]
	\item $\texttt{SEL}: (\mA^{\tuple{k \shortminus 1}}, \mX^{\tuple{k \shortminus 1}}) \mapsto \mS_k \in \sR^{N_k \times N_{k \shortminus 1}}$, defines how to aggregate the $N_{k \shortminus 1}$ nodes in the input graph into $N_k$ \textit{supernodes}.
    \item $\texttt{RED}: (\mX^{\tuple{k \shortminus 1}}, \mS_k) \mapsto \mX^{\tuple{k}} \in \sR^{N_k \times d_x}$, creates the supernode features by combining the features of the nodes assigned to the same supernode. A common way to implement \texttt{RED} is $\mX^{\tuple{k}} = \mS_k \mX^{\tuple{k \shortminus 1}}$.
	\item $\texttt{CON}: (\mA^{\tuple{k \shortminus 1}}, \mS_k) \mapsto \mA^{\tuple{k}} \in \sR^{N_k \times N_k}_{\geqzero}$, generates the edges (and, potentially, the edge features) by connecting the supernodes. A typical \texttt{CON} is $\mA^{\tuple{k}} = \mS_k \mA^{\tuple{k \shortminus 1}} \mS_k^{\top}$.
\end{itemize}
A fourth function is used to \textit{lift}, i.e., upsample, supernode features to the associated nodes in the original graph:
\begin{itemize}[leftmargin=1.5em, topsep=0pt]
    \item $\texttt{LFT}: (\mX^{\tuple{k}}, \mS_k) \mapsto \widetilde{\mX}^{\tuple{k \shortminus 1}} \in \sR^{N_{k \shortminus 1} \times d_x}$, can be implemented as $\widetilde{\mX}^{\tuple{k \shortminus 1}} = \mS^{+}_k\mX^{\tuple{k}}$, where $\mS_{k}^{+}$ is the pseudo-inverse of $\mS_{k}$.
\end{itemize}

As for \gls{tmp} and \gls{smp}, there is a strong analogy also between downsampling in time and space. Notably, the selection matrix $\mS_k$, hereinafter called \textit{spatial downsampling matrix}, plays the same role as $\mT_l$. Both matrices, indeed, reduce the input dimensionality conditioned on the underlying structure of the data, which in the temporal domain can be expressed as the path graph connecting the time steps. However, irregularities in arbitrary graphs make it challenging to define concepts like ``1-every-$k$''. An interpretation is given by the $k$-MIS  method~\cite{bacciu2023generalizing}, which relies on the concept of maximal $k$-independent sets to keep in the pooled graph the nodes that cover uniformly the $k$-th power graph.
Such a symmetry between the spatial and temporal operators will be pivotal for the design of our architecture.

\begin{figure*}[h]
    \centering
    \includegraphics[width=\textwidth]{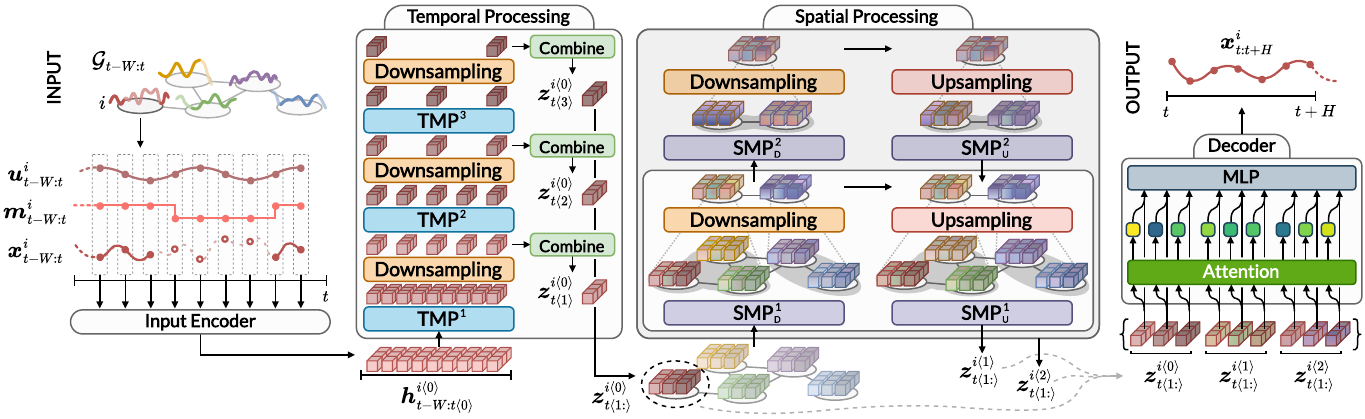}
    \caption{Overview of the proposed architecture. Given input data $\gG_{t-W:t}$, all information associated with every $i$-th node and time step $t$ is encoded in vectors $\smash{\vh^{i\tuple{0}}_{t\tuple{0}}}$, then processed node-wise along the temporal dimension by alternating \gls{tmp} and downsampling. After each $l$-th layer, the sequences are combined in a single vector $\vz^{i\tuple{0}}_{t\tuple{l}}$, which is then processed along the spatial dimension by alternating \gls{smp} and pooling. Representations at each $k$-th pooling layer are then recursively un-pooled up to the initial node level, obtaining $\smash{\vz^{i\tuple{k}}_{t\tuple{1:}}}$. Finally, the $L(K+1)$ encodings $\smash{\vz^{i\tuple{0:}}_{t\tuple{1:}}}$ are combined through an attention mechanism and fed to an \acrshort{mlp} to obtain the predictions.}
    \label{fig:architecture}
\end{figure*}

\subsection{Spatiotemporal Missing Data Distributions}
\label{sec:md_patterns}

Following previous works~\cite{yi2016stmvl, cini2022filling}, we categorize missing data patterns according to the conditional distribution $p\left(\vm^i_t\, |\, \mM_{\leq t} \right)$. We call \textit{point missing} the case where the probability $p\left(\vm^i_t = \vzero\right)$ of a datum being missing at a given node and time step is unconditioned and constant across nodes and time steps, i.e., 
\begin{equation}
    p\left(\vm^i_t\right) = \gB(1-\eta) \quad \forall\ i, t \label{eq:point}
\end{equation}
where $1-\eta$ is the mean of the Bernoulli distribution.
This setting is also known as \textit{general} missing~\cite{rubin1976inference}, as realizations of the mask have a haphazard pattern. 
Very often in \glspl{sn}, instead, $p\left(\vm^i_t\right)$ does depend on realizations of the missing data distribution at other nodes, time steps, or a combination of them (e.g., due to faults or blackouts). We refer to this setting as \textit{block missing} and decline it differently according to the dimensions of interest. 
In \textit{temporal block missing}, $p\left(\vm^i_t\right)$ depends on the realization of the missing data distribution at the previous time step, i.e., 
\begin{equation}
    p\left(\vm^i_t\, |\, \vm^i_{t-1}\right) \neq p\left(\vm^i_t\right) . \label{eq:block_t}
\end{equation}
Similarly, in \textit{spatial block missing}, $p\left(\vm^i_t\right)$ is conditioned on the simultaneous realizations at neighboring nodes, i.e., 
\begin{equation}
    p\left(\vm^i_t\, \big|\, \big\{\vm^j_t\big\}^{j \in \gN(i)}\right) \neq p\left(\vm^i_t\right) . \label{eq:block_s}
\end{equation}
The \textit{spatiotemporal block missing} combines~\autorefseq{eq:block_t}{eq:block_s} as:
\begin{equation}
    p\left(\vm^i_t\, \big|\, \vm^i_{t-1}, \big\{\vm^j_{t}\big\}^{j \in \gN(i)}\right) \neq p\left(\vm^i_t\right) . \label{eq:block_st}
\end{equation}
Note that in \autorefseq{eq:point}{eq:block_st} we considered a simplified case with a single channel in the observations, i.e., $d_x=1$. Instead, in the multivariate case $d_x>1$, we assume the missing data distribution of each channel to be independent of the others.

\section{Proposed Architecture}
\label{sec:architecture}

The dynamics in spatiotemporal data are governed by relationships spanning both the temporal and spatial dimensions. 
Operators that exploit the temporal or spatial structure underlying the data during processing, like those in~\autoref{sec:stmp}, capture such dynamics effectively.
Missing values, however, pose a serious challenge in identifying the correct dynamics. 
In addition, if the distribution of missing values is unknown, it is crucial to adaptively focus on different spatiotemporal scales conditioned on the input data. 
When a whole block of data is missing at a node (i.e., sensor), it could be beneficial to consider recent observations of the spatial neighbors. Conversely, when data are missing at neighboring nodes, data far back in time might be more informative.

In this section, we introduce \gls{model}, an architecture for graph-based forecasting of spatiotemporal data with missing values following arbitrary patterns. 
To learn spatiotemporal representations at different scales, we rely on hierarchical downsampling, which progressively reduces the size of the input and enables efficient learning of long-range dependencies.
In time, \gls{model} learns multiple -- yet limited -- representations from the input sequences, each at a different temporal scale. 
Similarly, we process the spatial dimension by propagating messages along a hierarchy of pre-computed coarsened graphs. 
Notably, our approach combines the advantages of the \gls{tas} and \gls{tts} paradigms, enabling adaptive expansion of the receptive field while keeping the computational and memory complexity under control.

The key components of \gls{model} are (1) an \textbf{input encoder}, (2) a \textbf{temporal processing} module, (3) a \textbf{spatial processing} module, and (4) an adaptive \textbf{decoder}. The first three blocks extract hierarchical representations of the input at different spatial and temporal scales, while the last block reweighs the representations and outputs the predictions.
The whole architecture is trained end-to-end to minimize the forecasting error over the prediction horizon~\autorefp{eq:loss}. Notably, our approach does not require missing data imputation as a pre-processing step. The latter might introduce a bias in the data and is generally cumbersome, as it turns the forecasting task into a two-step procedure. Also, domain knowledge is usually required to select the proper imputation technique.
In contrast, \gls{model} handles missing data by automatically reweighing the different spatiotemporal dynamics present in the data while learning to solve the downstream task.

\autoref{fig:architecture} shows an overview of the architecture. In the figure and the following, we use the superscript and subscript indices ${\,\cdot\,}^{\tuple{k}}_{\tuple{l}}$ to refer to the $k$-th spatial and $l$-th temporal scale, respectively, with $k=l=0$ corresponding to the input scales. 

\paragraph{Input encoder} %
The input encoder combines the information associated with the $i$-th node at generic time step $t$, i.e., observations $\vx^{i}_{t}$, exogenous variables $\vu^{i}_{t}$, and mask $\vm^{i}_{t}$ (see \autoref{sec:preliminaries}), in an embedding vector $\vh^{i\tuple{0}}_{t\tuple{0}} \in \sR^{d_h}$ with a \gls{mlp}:
\begin{equation}
    \vh^{i\tuple{0}}_{t\tuple{0}} = \textsc{MLP}\left(\vx^{i}_{t}, \vu^{i}_{t}, \vm^{i}_{t}, \vtheta^{i}\right). \label{eq:enc}
\end{equation}
Here, $d_h$ is the size of all latent representations learned within our model and $\vtheta^{i}$ is a vector of node-specific trainable parameters that facilitates node indetification~\cite{cini2023taming}. 
Missing values are imputed using the last observed value.
Obtained representations are then refined by downstream components exploiting temporal and spatial dependencies.

\paragraph{Temporal processing} %
The temporal processing module acts on $\vh^{i\tuple{0}}_{t \shortminus W:t\tuple{0}}$ to generate, for every $i$-th node, $L$ representations $\big\{\vz^{i\tuple{0}}_{t\tuple{l}}\big\}_{l=1,\dots,L}$, each associated with a different temporal scale. In particular, the $l$-th processing layer takes as input the sequence $\vh^{i\tuple{0}}_{t \shortminus W_{l \shortminus 1}:t\tuple{l \shortminus 1}}$ and outputs a new sequence of updated encodings decimated by a factor of $d$. Hence, the output sequence has length $W_l=\left\lceil\frac{W_{l \shortminus 1}}{d}\right\rceil$, with $W_0=W$ being the original time series' length.
We can write the operations performed by layer $l$ as
\begin{equation}
    \vh^{i\tuple{0}}_{t \shortminus W_{l}:t\tuple{l}} = \mT_l\left(\textsc{TMP}^{l}\Big( \vh^{i\tuple{0}}_{t \shortminus W_{l \shortminus 1}:t\tuple{l \shortminus 1}}\Big)\right),
\end{equation}
where $\textsc{TMP}^{l}$ is the \gls{tmp} operator defined in~\autoref{sec:stmp} acting at temporal scale $l - 1$ and $\mT_l \in \sR^{W_l \times W_{l \shortminus 1}}$ is the temporal downsampling matrix defined in~\autoref{sec:downsamplig}. The downsampling operation progressively expands the temporal receptive field, allowing us to capture increasingly slower temporal dynamics in the input.

We then combine the sequences $\smash{\vh^{i\tuple{0}}_{t \shortminus W_l:t\tuple{l}}}$ at each layer $l$ into a single representation $\vz^{i\tuple{0}}_{t\tuple{l}} \in \sR^{d_h}$, i.e.,
\begin{equation}
    \vz^{i\tuple{0}}_{t\tuple{l}} = \textsc{Combine} \left( \vh^{i\tuple{0}}_{t \shortminus W_{l}:t\tuple{l}} \right) .
\end{equation}
A simple implementation is to return the encoding associated with the last time step, i.e., $\vz^{i\tuple{0}}_{t\tuple{l}} = \vh^{i\tuple{0}}_{t-1\tuple{l}}$.

\begin{figure}[t]
    \begin{center}
    \includegraphics[width=\linewidth]{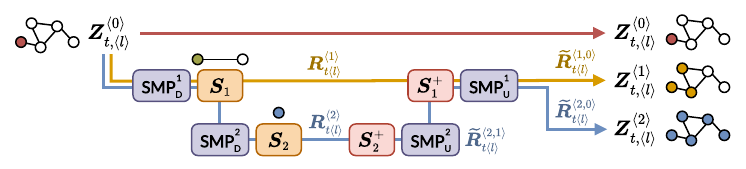}
    \caption{Details of the spatial processing procedure.}
    \label{fig:pooling_scheme}
    \end{center}
\end{figure}

\paragraph{Spatial processing} %
The set of multi-scale temporal encodings ${\big\{ \vz^{i\tuple{0}}_{t\tuple{l}} \big\}_{l=1,\dots,L}}$ is
processed spatially with an analogous hierarchical approach.
In particular, representations associated with the $k$-th spatial scale are obtained by applying $k$ times a combination of an \gls{smp} layer followed by the graph pooling operation described in~\autorefp{sec:downsamplig}:
\begin{equation}
    \mR^{\tuple{k}}_{t\tuple{l}} = \mS_{k} \left( \textsc{SMP}_{\text{D}}^{k}\left( \mR^{\tuple{k-1}}_{t\tuple{l}}, \mA^\tuple{k-1} \right) \right) , \label{eq:smp_pool}
\end{equation}
where $\mS_k \in \sR^{N_k \times N_{k-1}}$ is the spatial downsampling matrix generated by the pooling's \texttt{SEL} function, $\textsc{SMP}_{\text{D}}^{k}$ is the \gls{smp} operator applied at $k$-th layer before downsampling, $\mA^\tuple{0}=\mA$ and $\mR^{\tuple{0}}_{t\tuple{l}} = \mZ^{\tuple{0}}_{t\tuple{l}}$. 
At each spatial resolution level $k$, the encodings $\mR^{\tuple{k}}_{t\tuple{l}} \in \sR^{N_k \times d_h}$ are associated with supernodes rather than the nodes in the original graph.
Since the predictions must ultimately be computed for the nodes in the original graph, the coarsened spatial representations must be brought back to the initial spatial resolution. 
For this reason, we mirror the operations carried out in~\autoref{eq:smp_pool} and recursively lift supernodes' encodings back to the associated nodes in previous layers. Starting from the coarsened representation $\widetilde\mR^{\tuple{k, k}}_{t\tuple{l}} = \mR^{\tuple{k}}_{t\tuple{l}}$, the lifting step from layer $j$ to $j-1$ can be described as
\begin{equation}
    \widetilde\mR^{\tuple{k, j-1}}_{t\tuple{l}} = \textsc{SMP}_{\text{U}}^{j} \left(\mS_{j}^{+} \left( \widetilde\mR^{\tuple{k, j}}_{t\tuple{l}} \right), \mA^{\tuple{j-1}} \right) . \label{eq:smp_lift}
\end{equation}
After $k$ recursive application of~\autoref{eq:smp_lift}, the representations associated with the $k$-th spatial scale $\smash{\mR^{\tuple{k}}_{t\tuple{l}}}$ are propagated to the original graph and we assign $\mZ^{\tuple{k}}_{t\tuple{l}} = \widetilde\mR^{\tuple{k, 0}}_{t\tuple{l}}$. The details are depicted in~\autorefp{fig:pooling_scheme}.

\paragraph{Decoder} The $L(K+1)$ representations $\left\{\mZ^{\tuple{k}}_{t\tuple{l}}\right\}^{k=0,\dots, K}_{l=1,\dots, L}$ obtained in previous steps are associated with different temporal and spatial scales. To condition the importance of a representation to the prediction based  on the input dynamics and the missing data pattern, we first compute an adaptive weight $\alpha^{i\tuple{k}}_{t\tuple{l}} \in [0, 1]$ for each representation as
\begin{equation}
\label{eq:alphas}
    \left\{\alpha^{i\tuple{k}}_{t\tuple{l}}\right\} = \softmax \left\{\vz^{i\tuple{k}}_{t\tuple{l}} \Theta_{\alpha}\right\}^{k=0,\dots, K}_{l=1,\dots, L},
\end{equation}
with $\Theta_{\alpha} \in \sR^{d_h \times 1}$ being a matrix of learnable parameters. 
The weights can be interpreted as soft attention coefficients that select the appropriate spatiotemporal scale for computing the predictions.
As we show in~\autoref{sec:interpretability}, they offer a tool to analyze how the model focuses on different scales according to the dynamics and observability of the input.

The representations are then weighed and combined into a single final representation $\tilde\vz^i_t \in \sR^{d_h}$ as
\begin{equation}
    \tilde\vz^i_t = \sum_{l=1}^{L}\sum_{k=0}^{K} \alpha^{i\tuple{k}}_{t\tuple{l}}\vz^{i\tuple{k}}_{t\tuple{l}} .
\end{equation}
Finally, we use a multi-step \gls{mlp} decoder to obtain the predictions for each $i$-th node:
\begin{equation}
    \hat\vx^i_{t:t+H} = \mlp\left( \tilde\vz^i_t \right) .
\end{equation}

The pool of multi-scale representations and the adaptive reweighting enable more complex decoding strategies; we discuss them in~\appref{sec:model_appendix}.

\subsection{Implementation Details}
\label{sec:implementation}

While the proposed framework is general and different \gls{tmp}, \gls{smp}, and downsampling operators could be chosen, in this section we discuss the specific choices we made for our model.
As \gls{tmp}, we use \glspl{gru}~\cite{cho2014learning} combined with the standard $1$-every-$k$ temporal downsampling, which results in something similar to a dilated \gls{rnn}~\cite{chang2017dilated}. 
Referring to~\autoref{eq:smp}, we consider two variants with isotropic and anisotropic functions $\phi$ to compute the messages.
As isotropic \gls{smp}, we choose the diffusion-convolutional operator~\cite{atwood2016diffusion}, while for the anisotropic variant, we use the operator introduced by~\citet{cini2023taming}.
We use the sum as the aggregation function ($\textsc{Aggr}$) and we add aggregated messages to the source-node features $\vx^{i, l \shortminus 1}_{t}$ after an affine transformation (update function $\gamma$). We rely on the $k$-MIS pooling method (presented in~\autoref{sec:downsamplig}) to obtain spatial downsampling matrices $\mS_k \in \left[ 0, 1 \right]^{N_k \times N_{k \shortminus 1}}$ that assign each node to exactly one supernode. Notably, $k$-MIS is expressive, i.e., if two graphs are distinguishable, then the pooled graphs remain distinguishable~\cite{bianchi2023expressive}. 
For \texttt{RED}, \texttt{CON} and \texttt{LFT}, we use, respectively,
\begin{align*}
    \mX^{\tuple{k}} &= \mS_k \mX^{\tuple{k \shortminus 1}}, \\
    \mA^{\tuple{k}} &= \mS_k \mA^{\tuple{k \shortminus 1}} \mS_k^{\top}, \\
    \widetilde{\mX}^{\tuple{k \shortminus 1}} &= \mS^{+}_k\mX^{\tuple{k}}.
\end{align*}
More details on the implementation are in~\appref{sec:model_appendix}.

\paragraph{Analogy with filterbanks} The hierarchical processing scheme presented in~\autoref{sec:architecture} is inspired by the design of filterbanks in digital and graph signal processing~\cite{strang1996wavelets, tremblay2016subgraph, tremblay2018design}. 
Graph filterbanks often rely on specific node decimation operators that discard approximately half of the node at each application~\cite{shuman2015multiscale}.
Despite completely removing some nodes, when such techniques are paired with \gls{smp}, the original graph signal can be reconstructed without losses from the pooled one.
Such an approach has been adapted to the graph pooling framework~\cite{bianchi2019hierarchical} and represents an alternative to $k$-MIS pooling.

\subsection{Scalability}
\label{sec:scalability}

Maintaining multiple representations adds a computational burden to the already expensive operations performed by \glspl{stgnn} for standard spatial and temporal processing. 
For this reason, we adopted a series of architectural design choices to reduce the computational and memory complexity.
As previously discussed, we follow the \gls{tts} paradigm~\autorefp{sec:stmp}, which allows us to perform \gls{smp} operations on a single static graph, regardless of the length of the sequence being processed. 
Additionally, by integrating downsampling operations, the amount of computation reduces progressively with the layers, as the number of time steps and nodes is decimated.
Temporal processing, for instance, depends on the dilation factor chosen rather than the number of layers, scaling as $\gO\left(NW\left(d/(d-1) \right)\right)$ in contrast to $\gO(NWL)$.
Notably, this regularization has a positive impact also on the spatial and temporal receptive fields, which grow exponentially faster than standard \gls{tts} architectures, ceteris paribus.

The chosen $k$-MIS pooling operator belongs to the class of sparse and non-trainable pooling methods, which have a low memory footprint and do not require additional computation. A pooling operator is said to be \textit{sparse} if every node is assigned to at most one supernode. Additionally, pooling operators whose selection matrix $\mS_k$ is learned together with the model's weights are called \textit{trainable} (\textit{non-trainable} otherwise). The use of non-trainable methods allows us to compute the downsampling matrices once in a pre-processing step, making the spatial downsampling computationally comparable to the temporal one. While it can also act as a regularization for large models with high capacity, this lack of flexibility might introduce a too restrictive bias for its strong reliance on graph topology. In~\appref{sec:traffic_experiment}, we show an empirical example.

\section{Related Work}
\label{sec:rel_work}

Prominent representatives of \glspl{stgnn} for forecasting include
graph-based \glspl{rnn}, which integrate \gls{smp} within temporal processing~\cite{seo2018structured, li2018diffusion, zhang2018gaan, yu2019st, bai2020adaptive}, and convolutional approaches, which apply convolutions or attention mechanisms across temporal and spatial dimensions~\cite{yu2018spatio, wu2019graph, wu2022traversenet}.
Remarkably, none of these methods explicitly consider missing data in the input. 
GRIN~\cite{cini2022filling} and other \glspl{stgnn}~\cite{wu2021inductive, marisca2022learning} specifically address missing data in time series with relational information but focus on imputation. Forecasting with missing data, instead, is explored by a limited number of works in the context of traffic analytics~\cite{zhong2021heterogeneous, wang2023traffic}.

Numerous techniques exist for handling missing data, such as \gls{grud}~\cite{che2018recurrent} which simultaneously imputes missing values while processing the sequence, along with approaches for multivariate time classification~\cite{shukla2020multi}, anomaly detection~\cite{bianchi2019learning}, and forecasting~\cite{tang2020joint}. Yet, none of these methods explicitly model the spatial dependencies in the data.

Pyramidal approaches to process inputs with missing (or noisy) data have been extensively studied in graph~\cite{shuman2015multiscale, tremblay2016subgraph} and discrete-time~\cite{strang1996wavelets} signal processing.
The idea of using a hierarchy of either temporal or spatial representations has also been adopted by deep learning models for time series classification~\cite{cui2016multi}, forecasting~\cite{wang2022spatiotemporal, chen2023multi}, and forecast reconciliation~\cite{rangapuram2021end, rangapuram2023coherent}. In particular, \citet{cini2023graph_based} propose an approach for hierarchical time series forecasting where learnable graph pooling operations are used to cluster the time series.

\section{Experiments}
\label{sec:experiments}
In this section, we report the results of the empirical analysis of our approach in different synthetic and real-world settings. We use \gls{mae} as the figure of merit, averaged over only valid observations~\autorefp{eq:loss}. The code to reproduce the experiments and the instructions to download and pre-process the datasets are available online.\footnote{\url{https://github.com/marshka/hdtts}}

\paragraph{Baselines} We choose as baselines the following state-of-the-art \glspl{stgnn}:
\textbf{\gls{dcrnn}}~\cite{li2018diffusion}, a recurrent \gls{stgnn} using the diffusion-convolutional operator; \textbf{\gls{gwnet}}~\cite{wu2019graph},  a spatiotemporal convolutional residual network; \textbf{\gls{agcrn}}~\cite{bai2020adaptive}, an adaptive recurrent \gls{stgnn}; the four \textbf{\gls{ttsimp}}, \textbf{\gls{ttsamp}}, \textbf{\gls{tasimp}}, and \textbf{\gls{tasamp}} baselines introduced by~\citet{cini2023taming}, which are \gls{tts} and \gls{tas} architectures equipped with isotropic or anisotropic message passing, respectively. Given the similarity of \gls{ttsimp} and \gls{ttsamp} with our model, except for the hierarchical processing, they represent an ablation study about the effectiveness of learning hierarchical representations in \gls{tts} architectures.
Finally, we consider a \textbf{\acrshort{gru}} sharing the parameters across the time series. 
Since the baselines do not handle missing values in input, we impute them node-wise using the last observed value and concatenate the mask to the input as an exogenous variable.
In addition, we consider approaches that are specifically designed to process time series with missing data:
\textbf{\gls{grui}}, a \gls{gru} that imputes missing values with 1-step-ahead predictions during processing;
\textbf{\gls{grud}}~\cite{che2018recurrent};
 \textbf{\gls{grin}}, the Graph Recurrent Imputation Network~\cite{cini2022filling}, followed by a predictive decoder.
The details about the implementation of the baselines are in~\appref{sec:baselines_appendix}.

\paragraph{Missing data patterns}
Similar to previous works~\cite{yi2016stmvl, marisca2022learning}, we mask out observations in the data with missing data patterns that simulate realistic scenarios (see~\autoref{sec:md_patterns}). In the \textbf{\gls{point}} setting, each observation has a constant fixed probability $\eta$ of being missing~\autorefp{eq:point}. In the \textbf{\gls{block}} setting, in addition to the random point missing, we simulate sensor faults by letting sensor observations be missing for multiple consecutive steps with probability $p_f$~\autorefp{eq:block_t}. The duration of each fault is independently sampled from the same uniform distribution across sensors. Finally, \textbf{\gls{block_prop}} builds upon \gls{block} to reproduce faults affecting a group of sensors in a localized region and is obtained by propagating the fault to all $k$-hop neighbors with probabilities $\left[\, \vp_{g}\, \right]_k$~\autorefp{eq:block_st}. In all settings, we let the probabilities be the same for all sensors. Details on the distribution parameters can be found in~\appref{sec:simulated_masks}.

\begin{table}[t]
\caption{Forecasting error (MAE) on \gls{msods} with different missing data distributions. \textbf{Bold} formatting is used to mark best result in each setting. $^\dagger$Models without spatial message passing.}
\label{tab:mso_small}
\vskip 0.15in
\setlength{\tabcolsep}{5.5pt}
\setlength{\aboverulesep}{0pt}
\setlength{\belowrulesep}{0pt}
\renewcommand{\arraystretch}{1.2}
\begin{center}
\begin{small}
\begin{tabular}{l|c|c|c}
\toprule
\multicolumn{1}{c|}{\textbf{Model}}  & \textbf{\gls{point} (5\%)} & \textbf{\gls{block}} & \textbf{\gls{block_prop}} \\
\midrule
\gls{grud}$^\dagger$ & 0.385{{\tiny$\pm$0.012}} & 0.670{{\tiny$\pm$0.020}} & 1.081{{\tiny$\pm$0.003}} \\
\gls{grui}$^\dagger$ & 0.322{{\tiny$\pm$0.016}} & 0.619{{\tiny$\pm$0.011}} & 1.064{{\tiny$\pm$0.003}} \\
\gls{grin} & 0.163{{\tiny$\pm$0.008}} & 0.392{{\tiny$\pm$0.031}} & 0.895{{\tiny$\pm$0.012}} \\
\midrule
\gls{gru}$^\dagger$ & 0.346{{\tiny$\pm$0.027}} & 0.639{{\tiny$\pm$0.011}} & 1.137{{\tiny$\pm$0.008}} \\
\gls{dcrnn} & 0.291{{\tiny$\pm$0.277}} & 0.645{{\tiny$\pm$0.510}} & 1.103{{\tiny$\pm$0.001}} \\
\gls{agcrn} & 0.067{{\tiny$\pm$0.004}} & 0.366{{\tiny$\pm$0.013}} & 1.056{{\tiny$\pm$0.012}} \\
\gls{gwnet} & 0.089{{\tiny$\pm$0.002}} & 0.340{{\tiny$\pm$0.001}} & 0.955{{\tiny$\pm$0.012}} \\
\gls{tasimp} & 0.118{{\tiny$\pm$0.009}} & 0.323{{\tiny$\pm$0.011}} & 0.935{{\tiny$\pm$0.005}} \\
\gls{tasamp} & 0.063{{\tiny$\pm$0.003}} & 0.293{{\tiny$\pm$0.020}} & 0.868{{\tiny$\pm$0.006}} \\
\gls{ttsimp} & 0.113{{\tiny$\pm$0.008}} & 0.271{{\tiny$\pm$0.007}} & 0.697{{\tiny$\pm$0.005}} \\
\gls{ttsamp} & 0.096{{\tiny$\pm$0.004}} & 0.251{{\tiny$\pm$0.004}} & 0.669{{\tiny$\pm$0.013}} \\
\cmidrule[.8pt]{1-4}
\textbf{\gls{model}-IMP} & \textbf{0.058{{\tiny$\pm$0.004}}} & \textbf{0.247{{\tiny$\pm$0.002}}} & \textbf{0.651{{\tiny$\pm$0.023}}} \\
\textbf{\gls{model}-AMP} & 0.062{{\tiny$\pm$0.002}} & 0.261{{\tiny$\pm$0.009}} & 0.679{{\tiny$\pm$0.005}} \\
\bottomrule
\end{tabular}%
\end{small}
\end{center}
\vskip -0.1in
\end{table}

\begin{table*}[t]
\caption{Forecasting error (MAE) on real-world datasets with different missing data distributions. We use \textbf{bold} formatting to mark best results and N/A for runs exceeding resource capacity. $^\dagger$Models without spatial message passing.}
\vskip 0.1in
\label{tab:env}
\setlength{\tabcolsep}{5.5pt}
\setlength{\aboverulesep}{0pt}
\setlength{\belowrulesep}{0pt}
\renewcommand{\arraystretch}{1.1}
\begin{center}
\begin{small}
\begin{tabular}{l|cc|cc|cc|rr}
\cmidrule[.7pt]{2-9}
\multicolumn{1}{c}{} & \multicolumn{2}{c|}{\textsc{\gls{air}}} & \multicolumn{2}{c|}{\textsc{\gls{engrad}}} & \multicolumn{4}{c}{\textsc{\gls{pvus}}} \\
\toprule
\multicolumn{1}{c|}{\textbf{Model}} & \textbf{Original} & \textbf{+ Point} & \textbf{\gls{block}} & \textbf{\gls{block_prop}} & \textbf{\gls{block}} & \textbf{\gls{block_prop}} & \multicolumn{1}{c}{\textbf{Batch/s}} & \multicolumn{1}{c}{\textbf{GPU RAM}}\\
\midrule
\gls{grud}$^\dagger$ & 18.26{{\tiny$\pm$0.09}} & 19.23{{\tiny$\pm$0.08}} & 5.29{{\tiny$\pm$0.05}} & 5.41{{\tiny$\pm$0.01}} & 4.04{{\tiny$\pm$0.03}} & 4.27{{\tiny$\pm$0.01}}  & 5.07{\tiny$\pm$0.00} & 9.33 GB \\
\gls{grui}$^\dagger$ & 18.12{{\tiny$\pm$0.03}} & 19.07{{\tiny$\pm$0.01}} & 5.24{\tiny$\pm$0.04} & 5.39{\tiny$\pm$0.00} & 4.05{{\tiny$\pm$0.02}} & 4.29{{\tiny$\pm$0.02}} & 2.90{\tiny$\pm$0.00} & 10.28 GB \\
\gls{grin} & 16.85{{\tiny$\pm$0.05}} & 17.59{{\tiny$\pm$0.06}} & 4.91{{\tiny$\pm$0.04}} & 5.05{{\tiny$\pm$0.00}} & 3.62{\tiny$\pm$0.02} & 3.85{{\tiny$\pm$0.07}} & 1.52{\tiny$\pm$0.00} & 17.28 GB \\
\midrule
\gls{gru}$^\dagger$ & 18.17{{\tiny$\pm$0.03}} & 19.19{{\tiny$\pm$0.06}} & 5.30{{\tiny$\pm$0.03}} & 5.42{{\tiny$\pm$0.02}} & 3.98{{\tiny$\pm$0.02}} & 4.14{{\tiny$\pm$0.02}} & 11.59{\tiny$\pm$0.04} & 12.01 GB \\
\gls{dcrnn} & 16.99{{\tiny$\pm$0.09}} & 17.51{{\tiny$\pm$0.08}} & 5.14{{\tiny$\pm$0.06}} & 5.33{{\tiny$\pm$0.05}} & 3.54{{\tiny$\pm$0.01}} & 3.76{{\tiny$\pm$0.00}} & 1.36{\tiny$\pm$0.01} & 19.72 GB \\
\gls{agcrn} & 17.19{{\tiny$\pm$0.06}} & 17.92{{\tiny$\pm$0.05}} & 4.84{{\tiny$\pm$0.01}} & 5.10{{\tiny$\pm$0.06}} & 4.06{{\tiny$\pm$0.01}} & 4.20{{\tiny$\pm$0.04}} & 1.15{\tiny$\pm$0.01} & 23.40 GB \\
\gls{gwnet} & 15.89{{\tiny$\pm$0.04}} & 16.39{{\tiny$\pm$0.14}} & 4.59{{\tiny$\pm$0.04}} & 4.76{{\tiny$\pm$0.03}} & \textbf{3.48{{\tiny$\pm$0.05}}} & 3.71{{\tiny$\pm$0.03}} & 2.12{\tiny$\pm$0.00} & 16.02 GB \\
\gls{tasimp} & 16.54{{\tiny$\pm$0.03}} & 17.13{{\tiny$\pm$0.05}} & 4.98{{\tiny$\pm$0.01}} & 5.15{{\tiny$\pm$0.03}} & 3.60{{\tiny$\pm$0.02}} & 3.82{{\tiny$\pm$0.03}} & 2.68{\tiny$\pm$0.00} & 7.03 GB \\
\gls{tasamp} & 16.15{{\tiny$\pm$0.02}} & 16.58{{\tiny$\pm$0.10}} & 4.93{\tiny$\pm$0.02} & 5.11{\tiny$\pm$0.05} & N/A & N/A & \multicolumn{1}{c}{N/A} & \multicolumn{1}{c}{N/A} \\
\gls{ttsimp} & 16.25{{\tiny$\pm$0.01}} & 16.90{{\tiny$\pm$0.26}} & 4.81{{\tiny$\pm$0.07}} & 5.08{{\tiny$\pm$0.04}} & 3.50{{\tiny$\pm$0.01}} & 3.66{{\tiny$\pm$0.02}} & 18.84{\tiny$\pm$0.14} & 12.81 GB \\
\gls{ttsamp} & 15.63{{\tiny$\pm$0.06}} & 16.15{{\tiny$\pm$0.05}} & 4.70{{\tiny$\pm$0.00}} & 4.81{{\tiny$\pm$0.06}} & \textbf{3.46{{\tiny$\pm$0.03}}} & 3.65{{\tiny$\pm$0.05}} & 14.26{\tiny$\pm$0.08} & 12.81 GB \\
\cmidrule[.8pt]{1-9}
\textbf{\gls{model}-IMP} & 15.50{{\tiny$\pm$0.07}} & 15.94{{\tiny$\pm$0.10}} & \textbf{4.48{{\tiny$\pm$0.01}}} & \textbf{4.64{{\tiny$\pm$0.03}}} & \textbf{3.47{{\tiny$\pm$0.01}}} & \textbf{3.62{{\tiny$\pm$0.02}}} & 7.11{\tiny$\pm$0.03} & 10.86 GB \\
\textbf{\gls{model}-AMP} & \textbf{15.35{{\tiny$\pm$0.01}}} & \textbf{15.76{{\tiny$\pm$0.07}}} & 4.53{{\tiny$\pm$0.03}} & \textbf{4.65{{\tiny$\pm$0.04}}} & \textbf{3.47{{\tiny$\pm$0.02}}} & \textbf{3.61{{\tiny$\pm$0.02}}} & 6.21{\tiny$\pm$0.02} & 10.86 GB \\
\bottomrule
\end{tabular}%
\end{small}
\end{center}
\vskip -0.1in
\end{table*}

\subsection{\acrshort{mso}: a New Synthetic Benchmark}
\label{sec:mso}

We introduce a new synthetic dataset, called \textbf{\gls{mso}}, inspired by a popular benchmark in time series forecasting~\cite{bianchi2017recurrent}. Given a graph with binary adjacency matrix $\mA \in [0,1]^{N \times N}$, we first assign to all nodes sinusoids with incommensurable frequencies $\bar{\vx}^i_t = \sin{\left( \frac{t}{e^{i/N}}\right)}$ and then propagate the signals over the graph as
\begin{equation}
    \mX_{0:T} = \overline{\mX}_{0:T} + \dot{\mA}_{K}^{\top} \overline{\mX}_{0:T}, \label{eq:mso_signal}
\end{equation}
where $\dot{\mA}_{K}$ is a matrix where each column $i$ has only $5$ nonzero entries randomly selected from the $i$-th column of $\sum_{k=1}^{K}\mA^k$.
The resulting time series $\mX_{0:T}$ are aperiodic signals extremely difficult to predict unless the graph structure is used to recover the initial sinusoids. As such, \gls{mso} is a very effective benchmark to test the performance of spatiotemporal models (further details in~\appref{sec:app_mso}).
We consider three different settings with increasing difficulty levels: (1) \gls{point} missing with $\eta=0.05$; (2) \gls{block} missing with  $\eta=0.05$ and $p_f=0.01$ ($\approx 27\%$ data missing); (3) \gls{block_prop} missing with $\eta=0.05$, $p_f=0.005$, and $\vp_g = \left[\ 1\ \right]$, i.e., all faults being propagated to $1$-hop neighbors in $\dot{\mA}_{K}$ ($\approx 67\%$ data missing).

\autoref{tab:mso_small} shows the results of the analysis. Notably, \gls{model} outperforms all baselines in all settings.
The performance gap is notably pronounced when compared to more complex \glspl{stgnn} like \gls{gwnet} and \gls{agcrn}. These models, can~--~in principle~--~reach all the necessary spatial information through a single \gls{smp} step, as they learn a connectivity matrix.
Finally, the performance in the \gls{point} missing setting suggests that learning dedicated representations for the different spatiotemporal scales may also be beneficial when the missing information can be more easily recovered.

\begin{figure*}[t]
    \centering
    \includegraphics[width=\textwidth]{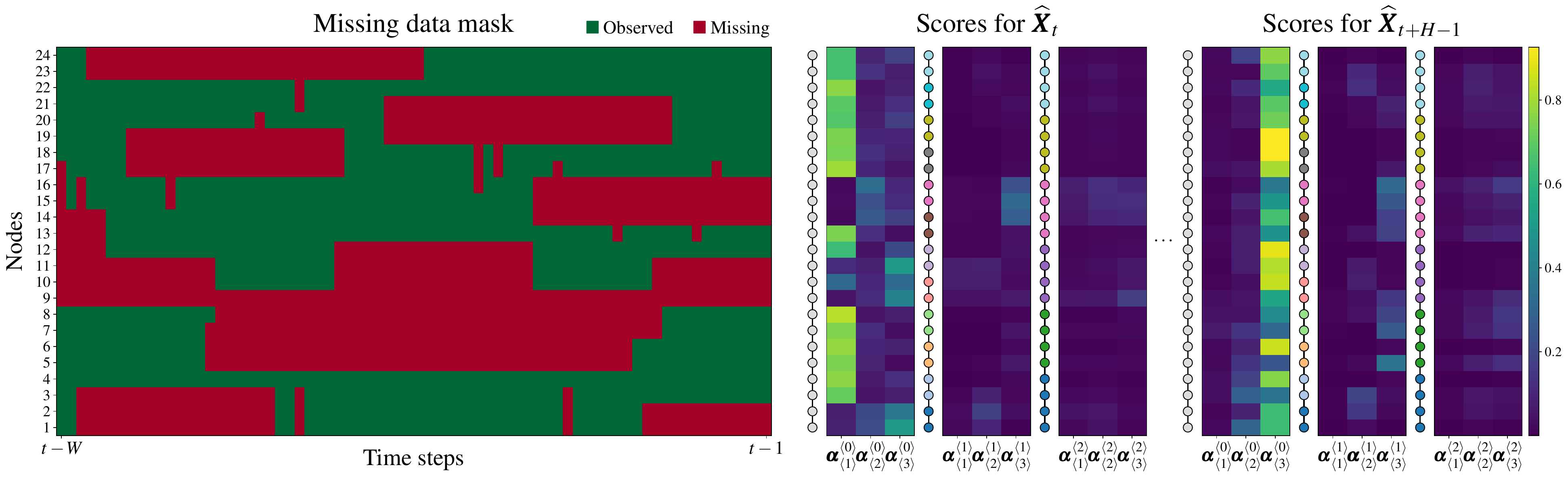}
    \caption{Decoder weights in \gls{msods}~(\gls{block_prop}). The graph used to produce this plot is the undirected line graph shown close to the scores associated with the first spatial resolution $k=0$. Node colors in the graphs associated with higher spatial scales show how nodes are clustered in supernodes by $\mS_k$ at each scale $k$.}
    \label{fig:mso_scores_paper}
\end{figure*}

\subsection{Real-world Benchmarks}
\label{sec:benchmarks}
In this experiment, we consider three real-world datasets for time series forecasting. 
\textbf{\gls{air}}~\cite{zheng2015forecasting} contains one year of hourly measurements from $437$ air quality monitoring stations in China. Since $25.67\%$ of the measurements are missing, it is a widely used benchmark for irregular spatiotemporal data analytics. \textbf{\gls{pvus}}~\cite{hummon2012sub} contains one year of solar power production from $5016$ simulated photovoltaic farms in the US with a $5$-minute sampling rate. We consider the subset of $1081$ western farms and aggregate measurements at 20-minute intervals. For the size of the graph, it is used to test the scalability of \glspl{stgnn}~\cite{cini2023scalable}. Finally, the second new dataset introduced in this paper, named \textbf{\gls{engrad}}, contains $3$ years of $5$ historical weather variables sampled hourly at $487$ grid points in England. The measurements are provided by \href{https://open-meteo.com/}{open-meteo.com}~\cite{Zippenfenig_Open-Meteo} and licensed under Attribution 4.0 International (CC BY 4.0). This dataset aims to fill a gap in the literature, by providing a spatiotemporal dataset with (1) multiple channels and (2) observations going beyond the slowest seasonality (i.e., one year). Following previous works, we obtain the adjacency matrices with a thresholded Gaussian kernel on the pairwise distances between sensors.
For \gls{air}, given the high number of missing values often occurring in blocks, we test the models on both the original data and in a \gls{point} setting with an additional $\eta=0.25$ portion of data masked out. For the other datasets, we consider the \gls{block} and \gls{block_prop} settings.
More details about the datasets and the missing data distributions can be found in~\appref{app:datasets}--\ref{sec:simulated_masks}.

\autoref{tab:env} shows the models' performance with different missing data patterns. Our approach ranks among the best-performing methods in all considered settings. Improvements are more evident with blocks of missing data spanning both time and space (\gls{block_prop}). Here, \gls{model} outperforms by a larger margin more complex \glspl{stgnn} and both IMP and AMP variants w.r.t.\ the respective \gls{tts} model.
This result underlines the advantage of combining hierarchical representations from multiple temporal and spatial scales when the amount of missing data is substantial. The results on \gls{air}, a dataset with a high ratio of missing observations, prove that \gls{model} is impactful in practical applications where missingness in data is a real concern. Finally, results on training speed and memory utilization on \gls{pvus} highlight the scalability advantages of our approach compared to state-of-the-art \glspl{stgnn}, which also produce higher forecasting errors. In~\appref{app:engrad_exp}, we report additional results for \gls{engrad} in a setting where some variables are considered exogenous rather than inputs and targets.

\subsection{Interpretability of Decoder Weights}
\label{sec:interpretability}
In~\autoref{fig:mso_scores_paper}, we show the missing data mask $\mM_{t-W:t}$ associated with the input and the attention scores computed by the decoder. The scores can be used to inspect which spatial and temporal scales the model focuses on when the pattern of missing data and the dynamics in the input change. For readability, we use an undirected line graph such that adjacent nodes are in consecutive positions in the figure. We consider the \gls{block_prop} setting and use the original adjacency matrix $\mA$ to propagate both the signal and the faults in~\autoref{eq:mso_signal}.
We use a multi-step decoder to compute different scores for the different forecasting steps (see~\appref{sec:model_appendix}); 
the figure shows only the scores associated with the first and last time step of the forecasts, i.e., $\widehat{\mX}_{t}$ and $\widehat{\mX}_{t+H-1}$. The scores computed for a time step are grouped into $K+1$ blocks, with $K=2$, each containing the scores of $L=3$ temporal scales associated with a given spatial scale.
As an example, the leftmost block contains the scores $\left[\bm{\alpha}_{\langle 1 \rangle}^{\langle 0 \rangle}, \bm{\alpha}_{\langle 2 \rangle}^{\langle 0 \rangle}, \bm{\alpha}_{\langle 3 \rangle}^{\langle 0 \rangle}\right]$ associated with the first spatial scale, i.e., the original graph.

It is possible to observe that, to predict the first step $\widehat{\mX}_{t}$, the model focuses on the first temporal scale if the most recent data is not missing. Instead, when data are missing at a given node $i$, more weight is given to higher levels in the spatial hierarchy. This indicates that the information is retrieved from the corresponding temporal scale of the $i$-th node's neighbors.
As expected, representations associated with slower dynamics -- in both time and space -- become more relevant when forecasting values farther in time. For example, we see that to predict $\widehat{\mX}_{t+H-1}$ the model focuses more on coarser temporal scales.
Also, note that the scores vary across nodes subject to the same or similar missing data patterns. Indeed, the most relevant spatiotemporal scales also depend on the specific dynamics of the time series associated with each node. In~\appref{app:decoder_weights}, we show additional analyses by considering a scenario where we keep the missing data patterns but assign each node the same signal.

\section{Conclusions}
\label{sec:conclusions}

We presented a novel framework for graph-based forecasting with missing data based on hierarchical spatiotemporal downsampling. Our model learns representations at different spatiotemporal scales, which are then combined by an interpretable attention mechanism to generate the forecasts. Thanks to this design, it can dynamically adapt the receptive field and handle different types of missing data patterns in a scalable fashion.
Empirical comparisons with state-of-the-art methods on both synthetic and real-world datasets showed notable performance improvements, both in terms of forecasting accuracy and computational efficiency. Improvements are even more evident in the challenging settings with blocks of missing values in space and time, where other \glspl{stgnn} models struggle.

The framework we proposed is very general and flexible.
In this work, we investigated a few options to implement the temporal and spatial components. However, different combinations and more powerful operators are likely to deliver further performance improvements.

\section*{Acknowledgments}
This work was supported by the Swiss National Science Foundation project FNS 204061: \emph{HigherOrder Relations and Dynamics in Graph Neural Networks}, the International Partnership Program of the Chinese Academy of Sciences under Grant 104GJHZ2022013GC, and by the Norwegian Research Council projects 345017 \emph{RELAY: Relational Deep Learning for Energy Analytics} and 300921 \emph{GraphDial: Graph-based Neural Models for Dialogue Management}.

\section*{Impact Statement}
This paper presents work whose goal is to advance the field of Machine Learning. There are many potential societal consequences of our work, none which we feel must be specifically highlighted here.


\bibliography{biblio}
\bibliographystyle{icml2024}


\newpage
\appendix
\onecolumn
\section*{Appendix}
\label{sec:appendix}
In this appendix, we provide further details about our architecture and additional material about the experimental evaluation.

\section{Datasets}
\label{app:datasets}
\begin{table*}[ht]
\caption{Statistics of the datasets and considered sliding-window parameters.}
\label{tab:datasets}
\vskip 0.1in
\setlength{\aboverulesep}{0pt}
\setlength{\belowrulesep}{0pt}
\renewcommand{\arraystretch}{1.25}
\centering
\begin{small}
\begin{tabular}{l|c c c c c c||c c}
\toprule
 \bfseries Datasets & \bfseries Type & \bfseries Nodes & \bfseries Edges & \bfseries Time steps & \bfseries Sampling Rate & \bfseries Channels & \bfseries Window & \bfseries Horizon\\
\toprule
\gls{msods} & Directed & 10,000 & 100 & 300 & N/A & 1 & 72 & 36 \\
\midrule
\gls{air} & Undirected & 437 & 2,730 & 8,760 & 1 hour & 1 & 24 & 6 \\
\gls{engrad} & Undirected & 487 & 2,297 & 26,304 & 1 hour & 5 & 24 & 6 \\
\gls{pvus} & Undirected & 1081 & 5,280 & 26,283 & 20 minutes & 1 & 72 & 6 \\
\bottomrule
\end{tabular}
\end{small}
\vskip -0.1in
\end{table*}

In this section, we report the details of the used datasets, with a particular focus on the two introduced in this paper: \gls{msods} and \gls{engrad}. Notably, all datasets used in our study are publicly available.

\subsection{\gls{msods}}
\label{sec:app_mso}
\begin{figure}[!ht]
    \centering
    \includegraphics[width=.8\linewidth]{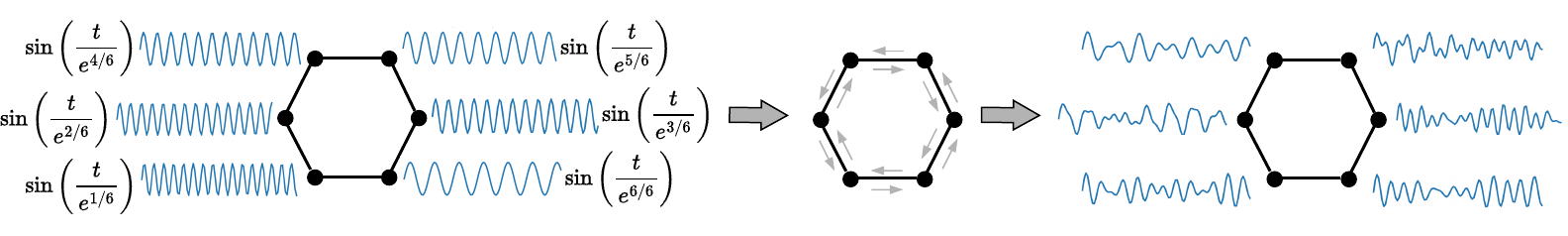}
    \caption{Schematic depiction of the generation of the \gls{msods} dataset.}
    \label{fig:mso_schema}
\end{figure}

\autoref{fig:mso_schema} illustrates the procedure for generating the \gls{msods} dataset. After creating a graph with a given topology, each node is assigned a sinusoid characterized by a frequency that is incommensurable with the ones at the other nodes. As such, summing one or more sinusoids results in a signal that is aperiodic and, thus, very difficult to predict. In addition, by aggregating from neighbors randomly chosen at different hops, predicting the signal becomes an even more challenging task.
To obtain such a signal, we combine each sinusoid with the sinusoids of the graph neighbors according to the propagation scheme described in~\autoref{sec:mso}.  For illustration purposes, in~\autoref{fig:mso_schema} we show a graph with a ring topology and propagate the sinusoids using the first-order adjacency matrix.

The details of the specific \gls{msods} dataset used in the paper are the following.
To build the adjacency matrix, we generated a random graph where each node has exactly 3 incoming edges.
The number of nodes in the graph is $N=100$ and the length of each time series is $10,000$ time steps.

\subsection{EngRAD}
\label{app:engrad}
The \gls{engrad} dataset contains measurements of $5$ different weather variables collected at $487$ grid points in England from 2018 to 2020. Data has been provided by open-meteo.com\footnote{\url{https://open-meteo.com/}}~\cite{Zippenfenig_Open-Meteo} and licensed under Attribution 4.0 International (CC BY 4.0). The numerical weather prediction model used to generate the data is ECMWF IFS, which has a spatial resolution of 9 km. The grid points are located in correspondence to cities. For each point, we provide basic information such as geographic coordinates, elevation, closest city, and the county to which it belongs. The physical variables collected are (1) air temperature at 2 meters above ground (°C); (2) relative humidity at 2 meters above ground ($\%$); (3) summation of total precipitation (rain, showers, snow) during the preceding hour (mm); (4) total cloud cover ($\%$); (5) global horizontal irradiance ($\text{W}/\text{m}^2$). The three complete years of observations allow for a comprehensive capture of phenomena associated with yearly seasonalities, a feature rarely found in existing benchmarks on spatiotemporal data, which are often too short.

\begin{figure}[ht]
     \centering
     \begin{subfigure}[t]{\textwidth}
         \centering
         \includegraphics[width=\textwidth]{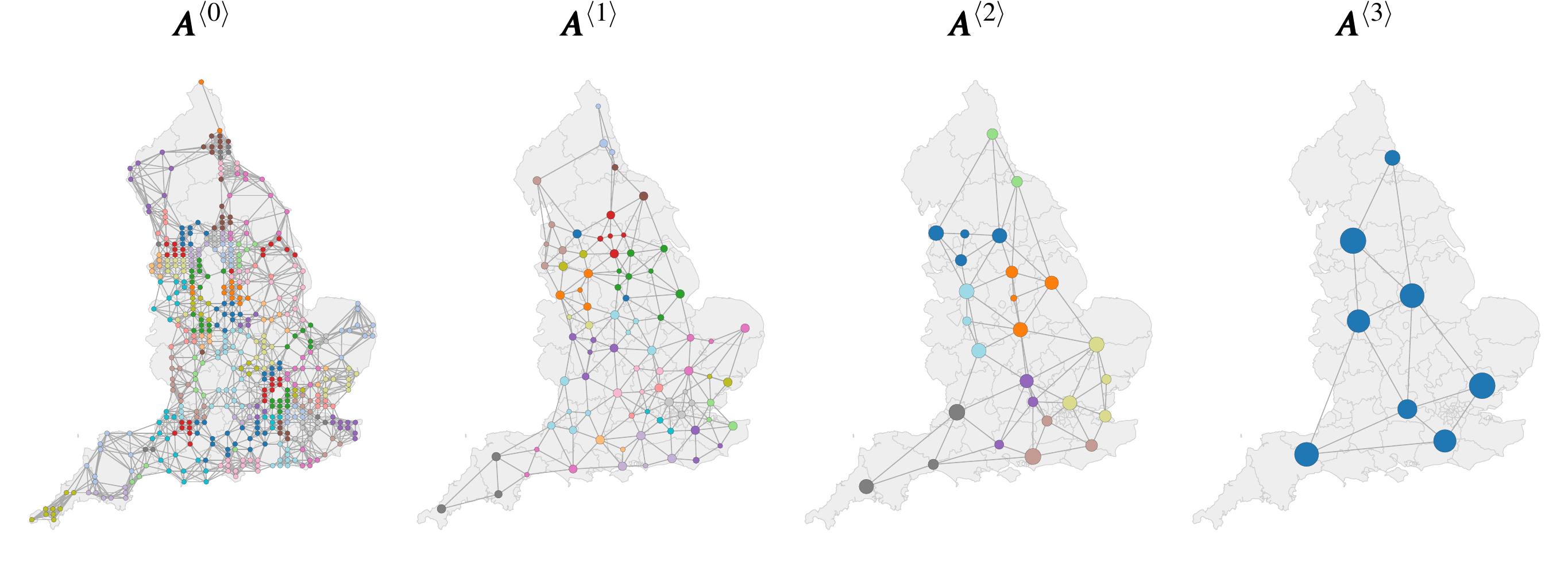}
        \caption{Plot of the graphs in geographic space. Nodes with the same color in a graph are assigned to the same supernode in the corresponding pooled graph (due to the limited amount of colors in the palette, the same color might be used in different clusters). The size of a supernode is proportional to its cardinality and its spatial coordinates are obtained by averaging the coordinates of the associated nodes with the \texttt{RED} operation.}
        \label{fig:engrad_pooled_graphs}
     \end{subfigure}
     \vskip0.5em
     \begin{subfigure}[t]{\textwidth}
         \centering
        \includegraphics[width=\textwidth]{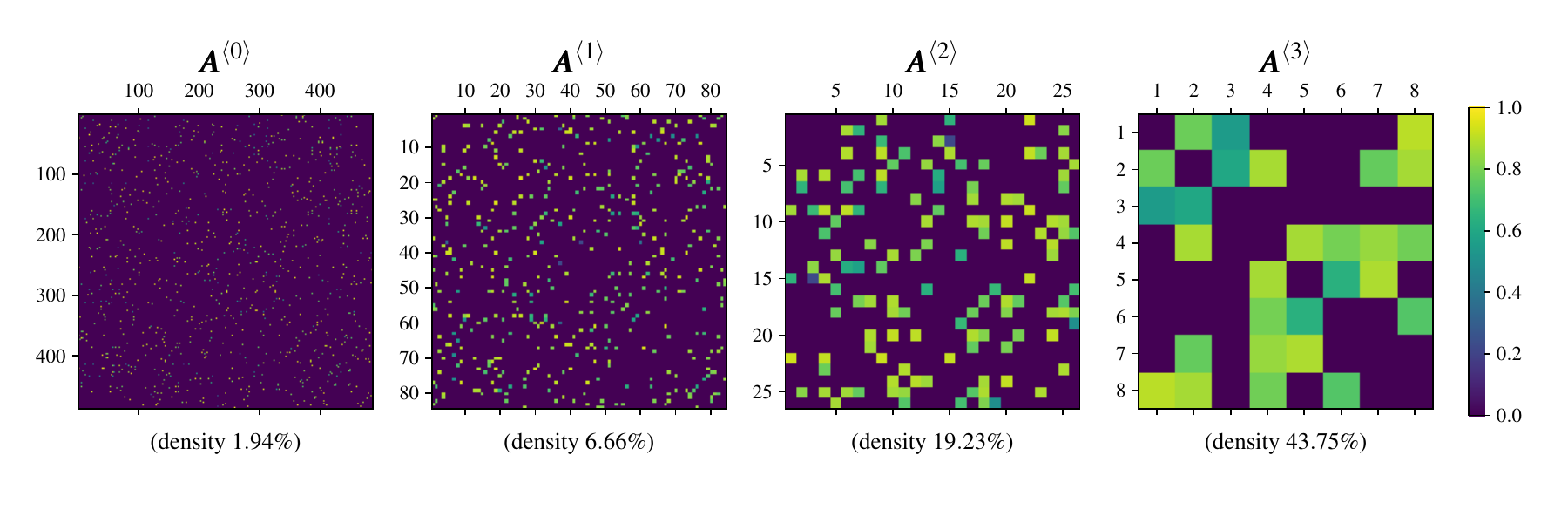}
        \vspace{-1.1cm}
        \caption{Adjacency matrices at different spatial scales.}
        \label{fig:engrad_pooled_adj}
     \end{subfigure}
    \vskip-0.5em
    \caption{Original and pooled graphs in \gls{engrad} obtained by recursive application of $k$-MIS pooling.}
    \label{fig:engrad_pooled}
\end{figure}

In~\autoref{fig:engrad_pooled}, we show a geographic map (\ref{fig:engrad_pooled_graphs}) and corresponding adjacency matrix (\ref{fig:engrad_pooled_adj}) of the \gls{engrad} graph ($\mA^{\tuple{0}}$) and the coarsened graphs ($\mA^{\tuple{1:4}}$) obtained with $k$-MIS pooling. As can be seen from the plots, the structural information of the original graph is preserved throughout the pooling operations.

\subsection{Other Datasets}
The two additional datasets considered in the experimental evaluation are \gls{air} and \gls{pvus}.
Introduced by~\citet{zheng2015forecasting}, \textbf{\gls{air}} contains hourly measurements of pollutant PM $2.5$ from $437$ air quality monitoring stations located in $43$ cities in China. Since $25.67\%$ of the measurements are missing, it is often used to evaluate imputation models~\cite{yi2016stmvl, cini2022filling, marisca2022learning}. The \textbf{\gls{pvus}} dataset~\cite{hummon2012sub} contains simulated energy production by $5016$ photovoltaic farms in the US over all the year $2006$. In the original datasets, samples are generated every $5$ minute. We use the data pre-processed by~\citet{cini2023scalable} and mask zero values associated with night hours (i.e., with no irradiance). We consider only the subset containing $1081$ farms in the western zones and aggregate observations at $20$-minute intervals by taking their mean.

\subsection{Data Pre-processing}
Following previous works~\cite{li2018diffusion, cini2023scalable}, to obtain the adjacency matrix in real-world datasets, we first compute a weighted, dense matrix containing the pairwise haversine distances between the geographic coordinates of the sensors. Then, we sparsify this matrix by applying a Gaussian kernel on the distances and zeroing out values under a threshold $\tau = 0.1$. For \gls{pvus} and \gls{engrad}, we additionally cap the maximum number of neighbors for each node by keeping the $8$ with the highest weight (i.e., shorter distance). To make the resulting graph undirected, we mirror edges that have a unique direction.

As these datasets contain communities of nodes very close in space and far from other communities, we insert additional edges to reduce the total number of connected components. In particular, an edge with minimum weight $\tau$ is added between the node in a disconnected community closest to the closest node in any other community. We repeat this step until the graph is connected or no edges can be added (e.g. if two connected components are very far apart).

For all datasets, we use as exogenous variables the encoding of the time of the day and the day of the year with two sinusoidal functions. For \gls{air}, we additionally include a one-hot encoding of the day of the week, as the presence of air pollutants may be influenced by humans' routines.

We split the datasets into windows of $W$ time steps, and train the models to predict the next $H$ observations. Values for $W$ and $H$ in each dataset are in~\autoref{tab:datasets}. For \gls{msods} and \gls{pvus}, we divide the obtained windows sequentially into $70\%/10\%/20\%$ splits for training, validation, and testing, respectively. For \gls{air}, we use as the test set the months of
March, June, September, and December, as done by~\citet{yi2016stmvl}. For \gls{engrad}, containing $3$ years of data, we use the year 2020 for testing and one week per month in the year 2019 for validation. 
We use all the samples that do not overlap with the validation and test sets to train the models.

For \gls{air} and \gls{mso}, we transform the data to have zero mean and unitary variance, while we rescale them to the $[0, 1]$ interval in \gls{engrad} and \gls{pvus}. The statistics used to transform the data are computed using only valid values in the training set.

\section{Experimental Setting}
\label{sec:experimental_setting}

\paragraph{Software \& Hardware}
\label{sec:software_harware}
All the code used for the experiments has been developed with Python~\cite{rossum2009python} and relies on the following open-source libraries: PyTorch~\cite{paske2019pytorch}; PyTorch Geometric~\cite{fey2019fast}; Torch Spatiotemporal~\cite{Cini_Torch_Spatiotemporal_2022}; PyTorch Lightning~\cite{Falcon_PyTorch_Lightning_2019}; Hydra~\cite{Yadan2019Hydra}; Numpy~\cite{harris2020array}; Scikit-learn~\cite{pedregosa2011scikit}.

Experiments were run on different workstations with varying models of processors and GPUs. However, we used the same workstation for measuring time and memory requirements in \gls{pvus}, whose results are reported in~\autoref{tab:env}. More details are provided later in this section.

\paragraph{Baselines}
\label{sec:baselines_appendix}
In the following, we report the hyperparameters used in the experiment for the considered baselines. Whenever possible, we relied on code provided by the authors or available within open-source libraries to implement the baselines.
To ensure a fair comparison, we tried to keep the number of trainable parameters of our model and each baseline in the same range.

\begin{description}[leftmargin=1em, itemindent=0em, itemsep=-.15em, topsep=-0.2em]
    \item[\textbf{\gls{dcrnn}}] We used the same parameters of the original paper~\cite{li2018diffusion}, with an embedding size of $64$ and a $K=2$ spatial receptive field. Following previous works~\cite{cini2023scalable}, we used an \gls{mlp} as the decoder with one hidden layer with $128$ units.
    \item[\textbf{\gls{gwnet}}] We used the same parameters reported in the original paper~\cite{wu2019graph}, except for those controlling the receptive field. Being \gls{gwnet} a convolutional architecture, this was done to ensure that the receptive field covers the whole input sequence.
    In particular, we used $8$ layers with temporal kernel size and dilation of $3$ when the input window is $24$ and $7$ layers with dilation and kernel size of $5$ when the input window is $72$.
    \item[\textbf{\gls{agcrn}}] We used the same hyperparameters reported in the original paper~\cite{bai2020adaptive}.
    \item[\textbf{\gls{grin}}] To implement this baseline, we used the unidirectional layer available in the Torch Spatiotemporal library\footnote{\url{https://github.com/TorchSpatiotemporal/tsl}} followed by an \gls{mlp} as the decoder. We kept the same hyperparameters of the original paper~\cite{cini2022filling}.
\end{description}
For the \glspl{gru}, namely \textbf{\gls{gru}}, \textbf{\gls{grud}}, and \textbf{\gls{grui}}, we used a size of $d_h=64$ for the hidden dimension, $L=2$ recurrent layers and one hidden layer with $128$ units for the \gls{mlp} used as decoder.
For the four \textbf{\gls{ttsimp}}, \textbf{\gls{ttsamp}}, \textbf{\gls{tasimp}}, and \textbf{\gls{tasamp}} baselines, we used the same hyperparameters reported in the paper~\cite{cini2023taming} and increased the \gls{smp} layers to $K=4$ to obtain a wider receptive field.

\paragraph{Training setting} We used AdamW~\cite{loshchilov2018decoupled} as the optimizer with an initial learning rate of $0.001$. 
We used the \texttt{ReduceLROnPlateau} scheduler of Pytorch that reduces the learning rate %
by a factor of $0.5$ if no improvements are noticed after $10$ epochs. We trained all models for $200$ epochs of $300$ randomly drawn mini-batches of $32$ examples and stopped the training if the \gls{mae} computed on the validation set did not decrease after $30$ epochs. We then used the weights of the best-performing model for evaluation on the test set. For the baselines that jointly perform imputation and forecasting, we used the imputation error as an auxiliary loss.

\paragraph{Simulated missing data patterns}
\label{sec:simulated_masks}
To evaluate the models' performance in function of the missing data pattern in input, we design a strategy to inject missing values in a controlled fashion. Given a sequence of observations of length $T$, we generate the missing data mask $\mM_{t:t+T}$ according to three parameters:
\begin{description}
    \item[\normalfont $\eta$:] The parameter of the Bernoulli distribution associated with the probability of an observation being missing.
    \item[\normalfont $p_f$:] The parameter of the Bernoulli distribution associated with the probability of a sensor fault (consecutive missing values).
        \begin{description}
            \item[\normalfont $s_{\min}, s_{\max}$:] The boundaries of the uniform distribution $\gU(s_{\min}, s_{\max})$ from which the length of the fault $s$ is sampled.
        \end{description}
    \item[\normalfont $\vp_g$:] The vector of the Bernoulli distributions' parameters associated with the probability of a fault to be propagated at neighboring nodes. The $k$-th element of the vector is associated with the $k$-th neighborhood order.
\end{description}
All parameters are constant across nodes and time steps, meaning that the process generating the missing data is stationary. For each dataset, we fix the seed used to generate the mask, ensuring that all models observe the same data in all runs. In the case of multivariate sensor observations, such as in \gls{engrad}, the mask is sampled independently across the channels, possibly resulting in partially available observations. We report the parameters used to generate the mask in the different settings in~\autoref{tab:mask_parameters}.
Note that in \gls{msods} even if we have $\vp_g = [1]$, the propagation is done with the matrix $\dot{\mA}_{K}$ defined in~\autoref{sec:mso}, which contains neighbors sampled from all orders up to $K$.
\begin{table}[t]
    \caption{Parameters used to generate the missing data mask in the different settings.}
    \label{tab:mask_parameters}
    \vskip 0.15in
    \setlength{\tabcolsep}{5.5pt}
    \setlength{\aboverulesep}{0pt}
    \setlength{\belowrulesep}{0pt}
    \renewcommand{\arraystretch}{1.2}
    \begin{center}
    \begin{small}
    \begin{tabular}{c|ccc|c|cc}
        \cmidrule[.9pt]{2-7}
          \multicolumn{1}{c}{} & \multicolumn{3}{c|}{\textbf{\gls{msods}}} & \textbf{\gls{air}} & \multicolumn{2}{c}{\textbf{\gls{engrad} / \gls{pvus}}} \\
          \toprule
         \textbf{Parameter} & \textbf{\gls{point}} & \textbf{\gls{block}} & \textbf{\gls{block_prop}} & \textbf{\gls{point}} & \textbf{\gls{block}} & \textbf{\gls{block_prop}} \\
         \midrule
         $\eta$ & $0.05$ & $0.05$ & $0.05$ & $0.25$ & $0.05$ & $0.075$ \\
         $p_f$ & $0$ & $0.01$ & $0.005$ & 0 & $0.01$ & $0.003$ \\
         $s_{\min}$ & -- & $8$ & $8$ & -- & $4$ & $6$ \\
         $s_{\max}$ & -- & $48$ & $48$ & -- & $12$ & $21$ \\
         $\vp_g$ & $0$ & $0$ & $[\begin{array}{c}1\end{array}]$ & $0$ & $0$ & $[\begin{array}{ccc}0.33 & 0.15 & 0.05\end{array}]$\\
         \bottomrule
    \end{tabular}
    \end{small}
    \end{center}
    \vskip -0.1in
\end{table}

\begin{figure}[ht]
    \centering
    \includegraphics[width=\textwidth]{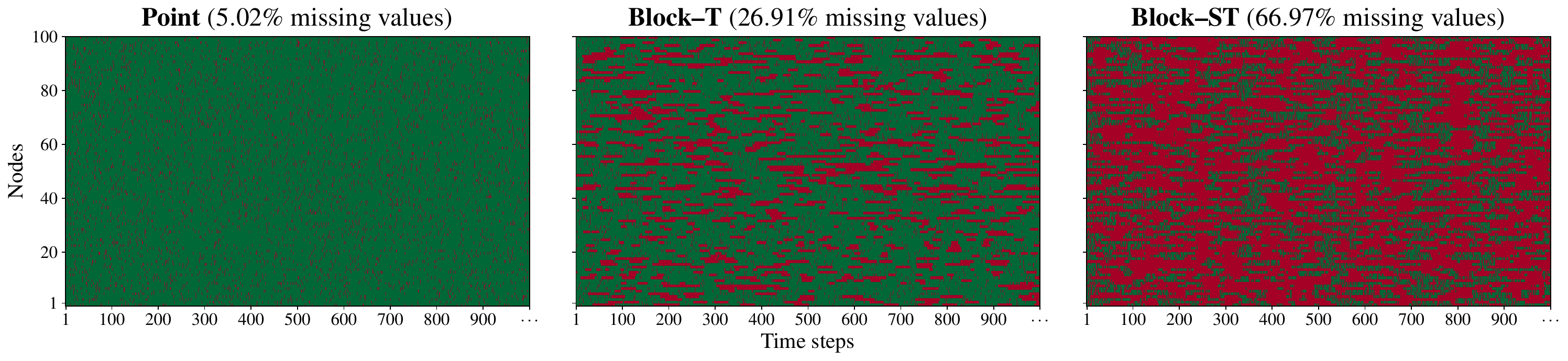}
    \caption{From left to right, the masks used in \gls{msods}. Green and red colors denote valid and missing observations, respectively.}
    \label{fig:masks_mso}
\end{figure}

To give an idea of how the missing data distributions look like, in~\autoref{fig:masks_mso} we report the missing data masks used in \gls{msods} for the three settings specified in~\autoref{tab:mask_parameters}.

During training, the generated mask is used to remove observations in both the input and ground-truth data. Conversely, when testing the models only the input is masked out and the forecasting error metrics are computed on all valid observations (i.e., not missing in the original dataset).

\paragraph{Time and memory requirements}
\label{sec:scalability_results}
In~\autoref{tab:env} of the paper, we report the time required for a single model update (expressed in batches per second) and the corresponding GPU memory usage for each method under consideration. To ensure a fair assessment, we record the time elapsed between the initiation of the inference step and the completion of the weights' update for a total of $50$ mini-batches of $16$ samples each, excluding the initial 5 and final 5 measurements to mitigate potential overhead impacts.
To measure the GPU memory reserved by each model, we relied on the NVIDIA System Management Interface\footnote{\url{https://developer.nvidia.com/nvidia-system-management-interface}}, which provides near real-time GPU usage monitoring.
We ran all experiments sequentially on a workstation running Ubuntu 20.04.6 LTS and equipped with one AMD Ryzen 9 5900X 12-core processor, 128GB 3200MHz DDR4 RAM, and two NVIDIA RTX A6000 GPU with 48 GB GDDR6 RAM.

\section{Architecture Details}
\label{sec:model_appendix}

In this section, we provide additional details about the implementation of our model.

\paragraph{Spatial message passing} We consider two different \gls{smp} operators, belonging to the isotropic and anisotropic categories, respectively. As isotropic \gls{smp}, we use the diffusion-convolutional operator~\cite{atwood2016diffusion}, that computes the messages for each $p$-hop neighborhood as
\begin{equation}
    \phi^{l} (\vx^{j,l \shortminus 1}_{t}, \tilde{a}_p^{ji}) = \tilde{a}_p^{ji} \vx^{j,l \shortminus 1}_{t}\Theta^l_{p} \label{eq:smp_imp}
\end{equation}
where $\phi^{l}$ is the message function~\autorefp{sec:stmp}, $\tilde{a}_p^{ji}$ is the weight of the edge between nodes $j$ and $i$ in $\mA^p$ normalized by the sum of the incoming edges' weight. For directed graphs, we further compute messages from $p$-hop neighbors in $\mA^{\top}$ with different parameters $\Theta^{\prime l}_{p}$, following previous works~\cite{li2018diffusion}. 

For the anisotropic variant, we rely on the anisotropic \gls{smp} layer introduced by~\citet{cini2023taming}, that computes the messages for each $p$-hop neighborhood as
\begin{gather}
    \vm^{ji, l \shortminus 1}_t = \xi\Big(\Big[\vx^{i, l \shortminus 1}_{t}\, \|\, \vx^{j, l \shortminus 1}_{t}\, \|\, a_{ji}\Big]\Theta_1^l\Big)\Theta_2^l , \label{e:amp_first} \\
    \tilde{a}^{ji, l \shortminus 1}_t = \sigma\Big(\vm^{ji, l \shortminus 1}_t \Theta_3^l\Big),\\
    \phi^{l} (\vm^{ji, l \shortminus 1}_t, \tilde{a}^{ji, l \shortminus 1}_t) = \tilde{a}^{ji, l \shortminus 1}_t \vm^{ji, l \shortminus 1}_t \label{eq:smp_amp}
\end{gather}
where $\Theta_1^l, \Theta_2^l$, and $\Theta_3^l$ are matrices of learnable parameters, $\|$ the concatenation operation, and $\sigma$ and $\xi$ the sigmoid and ELU activation functions, respectively.

These two strategies are only used to implement \gls{smp} during the downsampling stage, i.e., $\textsc{SMP}_{\text{D}}$.
To reduce the total number of parameters and operations, after upsampling we only perform a propagation step along the graph, i.e., we implement $\textsc{SMP}_{\text{U}}(\mX, \mA) = \mA^\top\mX$.

\paragraph{$k$-MIS pooling}
$k$-MIS pooling~\cite{bacciu2023generalizing} is a graph-based equivalent of evenly-spaced coarsening mechanisms for regular data. $k$-MIS pooling provides theoretical guarantees for distortion bounds on path lengths and the ability to preserve key topological properties in the coarsened graphs.
The method utilizes maximal $k$-independent sets to find a set of nodes that are approximately equally spaced in the original graph. The selected nodes are then used as vertices of the reduced graph. 
$k$-MIS pooling first identifies the supernodes with the centroids of the maximal $k$-independent set of a graph.
The centroids are selected with a greedy approach based on a ranking vector $\pi$ applied to the vertex features to speed up computation.
Depending on how the ranking $\pi$ is computed, $k$-MIS pooling declines as a trainable or non-trainable pooling operator.
In the former case, ranking is computed as $\pi = \mX^{\tuple{k \shortminus 1}}\mathbf{p}^{\top}$ where the projector $\mathbf{p}$ is the output of a trainable function.
The non-trainable version of $k$-MIS pooling, instead, relies on a pre-defined ranking function that does not depend on trainable parameters or node features. 
The latter is the setting adopted in this work, as we let the ranking be constant across the node, i.e., $\pi = \vone$. 
In addition, we convert directed graphs to undirected, as the method acts on undirected graphs.

\paragraph{Multi-step decoder} In some applications, the different dynamics captured by the multi-scale representations might be more or less relevant to the prediction depending on the distance of the target time step from the last observations. The structured multi-scale encodings, paired with the adaptive weighting mechanism, allow us to compute $H$ different score sets $\left\{\alpha^{i\tuple{k}}_{t+h\tuple{l}}\right\}_{h=0,\dots,H-1}$, and, hence, $H$ aggregated representations $\left\{\tilde\vz^i_{t+h}\right\}_{h=0,\dots,H-1}$, per node. The predictions are then obtained, $\forall\ h \in [0, H)$, as
\begin{equation}
    \hat\vx^i_{t+h} = \mlp\left( \tilde\vz^i_{t+h} \right) .
\end{equation}
We will use this implementation in~\appref{app:decoder_weights} to inspect the differences in the scores computed for the first and last time steps in the forecasting horizon.

\paragraph{Hyperparameters} We use an embedding size of $d_h=64$ for all hidden representations $\vh^i_t$ and $\vz^i_t$. In the input encoder, we concatenate the input with node embeddings $\vtheta^i$ of size $32$ and apply an affine transformation. For temporal processing, we use $L=4$ layers with a downsampling factor $d=3$. For spatial processing, we use $K=3$ pooling layers. The decoder is an \gls{mlp} with $2$ hidden layers with $128$ units. We use ELU as the activation function within the architecture. In the real-world datasets, we compute different attention weights for each forecasting step, as discussed in the previous section.

\begin{figure}[t]
     \centering
     \begin{subfigure}[t]{\textwidth}
         \centering
         \includegraphics[width=\textwidth]{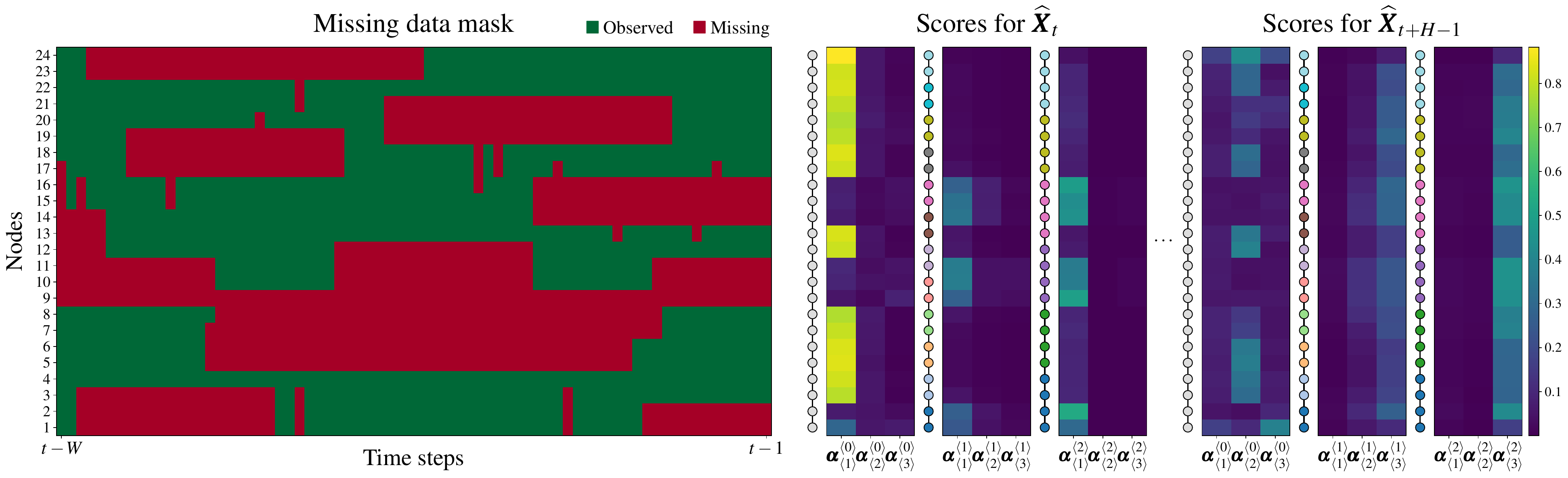}
         \caption{\gls{msods} dataset with the same time series associated with each node.}
         \label{fig:scores_lor}
     \end{subfigure}
     \vskip0.5em
     \begin{subfigure}[t]{\textwidth}
         \centering
         \includegraphics[width=\textwidth]{figs/scores_sinusoid.pdf}
         \caption{\gls{msods} dataset with the original data-generating process (different time series across nodes).}
         \label{fig:scores_sin}
     \end{subfigure}
    \caption{Decoder weights in \gls{msods}~(\gls{block_prop}) with the same missing data pattern but different node signals as input. The graph used to produce this plot is the undirected line graph shown close to the scores associated with the first spatial resolution $k=0$. Node colors in the graphs associated with higher spatial scales show how nodes are clustered in supernodes by $\mS_k$ at each scale $k$.}
    \label{fig:mso_scores}
\end{figure}

\section{Interpretability of Decoder Weights}
\label{app:decoder_weights}
In this appendix, we extend the analysis on the decoder weights reported in~\autoref{sec:interpretability} including a variant of \gls{msods} with the same time series associated with each node. Indeed, the scores are also influenced by the actual dynamics in the input time series, which can make it harder to see the relationship between the missing data patterns and the scores.
In~\autoref{fig:scores_lor}, we show the scores computed by the decode in this setting; we include the original plot of~\autoref{fig:mso_scores_paper} as~\autoref{fig:scores_sin} for a better comparison. In both settings, we can observe that missing data affecting a given spatiotemporal scale in a node results in a lower weight given to the corresponding representations. This is even more evident in the variant with the same time series in each node, where the contrast between high and low scores is more pronounced.

\section{\gls{engrad} Additional Results}
\label{app:engrad_exp}

\begin{table*}[t]
\caption{Forecasting error (MAE) on \gls{engrad} with different target variables and missing data distributions. We use \textbf{bold} formatting to mark best results. $^\dagger$Models without spatial message passing.}
\vskip 0.1in
\label{tab:engrad_exp}
\setlength{\tabcolsep}{5.5pt}
\setlength{\aboverulesep}{0pt}
\setlength{\belowrulesep}{0pt}
\renewcommand{\arraystretch}{1.1}
\begin{center}
\begin{small}
\begin{tabular}{l|cc|cc}
\cmidrule[.7pt]{2-5}
\multicolumn{1}{c}{} & \multicolumn{2}{c|}{\textsc{All variables}} & \multicolumn{2}{c}{\textsc{Temp. \& Rad.}} \\
\toprule
\multicolumn{1}{c|}{\textbf{Model}} & \textbf{\gls{block}} & \textbf{\gls{block_prop}} & \textbf{\gls{block}} & \textbf{\gls{block_prop}} \\
\midrule
\gls{grud}$^\dagger$ & 5.77{{\tiny$\pm$0.03}} & 6.01{{\tiny$\pm$0.06}} & 16.63{{\tiny$\pm$0.12}} & 16.82{{\tiny$\pm$0.12}} \\
\gls{grui}$^\dagger$ & 5.79{{\tiny$\pm$0.01}} & 5.95{{\tiny$\pm$0.01}} & 16.66{{\tiny$\pm$0.07}} & 16.80{{\tiny$\pm$0.04}} \\
\gls{grin} & 5.41{{\tiny$\pm$0.06}} & 5.61{{\tiny$\pm$0.03}} & 15.59{{\tiny$\pm$0.14}} & 15.80{{\tiny$\pm$0.11}} \\
\midrule
\gls{gru}$^\dagger$ & 5.84{{\tiny$\pm$0.00}} & 6.03{{\tiny$\pm$0.04}} & 16.77{{\tiny$\pm$0.08}} & 16.97{{\tiny$\pm$0.10}} \\
\gls{dcrnn} & 5.63{{\tiny$\pm$0.03}} & 5.90{{\tiny$\pm$0.04}} & 16.27{{\tiny$\pm$0.18}} & 16.52{{\tiny$\pm$0.18}} \\
\gls{agcrn} & 5.26{{\tiny$\pm$0.01}} & 5.53{{\tiny$\pm$0.02}} & 15.31{{\tiny$\pm$0.04}} & 15.71{{\tiny$\pm$0.09}} \\
\gls{gwnet} & 5.05{{\tiny$\pm$0.08}} & 5.22{{\tiny$\pm$0.07}} & 14.49{{\tiny$\pm$0.15}} & 14.87{{\tiny$\pm$0.07}} \\
\gls{tasimp} & 5.47{{\tiny$\pm$0.02}} & 5.67{{\tiny$\pm$0.05}} & 15.85{{\tiny$\pm$0.13}} & 16.13{{\tiny$\pm$0.16}} \\
\gls{tasamp} & 5.37{{\tiny$\pm$0.03}} & 5.58{{\tiny$\pm$0.09}} & 15.82{{\tiny$\pm$0.17}} & 15.86{{\tiny$\pm$0.10}} \\
\gls{ttsimp} & 5.21{{\tiny$\pm$0.04}} & 5.46{{\tiny$\pm$0.07}} & 15.32{{\tiny$\pm$0.18}} & 15.19{{\tiny$\pm$0.23}} \\
\gls{ttsamp} & 5.10{{\tiny$\pm$0.03}} & 5.31{{\tiny$\pm$0.04}} & 14.89{{\tiny$\pm$0.19}} & 15.02{{\tiny$\pm$0.07}} \\
\cmidrule[.8pt]{1-5}
\textbf{\gls{model}-IMP} & \textbf{4.93{{\tiny$\pm$0.02}}} & \textbf{5.12{{\tiny$\pm$0.03}}} & \textbf{14.34{{\tiny$\pm$0.09}}} & \textbf{14.41{{\tiny$\pm$0.07}}} \\
\textbf{\gls{model}-AMP} & \textbf{4.93{{\tiny$\pm$0.03}}} & \textbf{5.10{{\tiny$\pm$0.01}}} & \textbf{14.22{{\tiny$\pm$0.08}}} & \textbf{14.46{{\tiny$\pm$0.14}}} \\
\bottomrule
\end{tabular}%
\end{small}
\end{center}
\vskip -0.1in
\end{table*}

In this appendix, we consider two additional settings for the \gls{engrad} dataset. The first setting is similar to the one considered in~\autoref{sec:benchmarks}, but we do not compute loss and metrics on the prediction of the global horizontal irradiance variable during night hours when it is null.
In the second setting, we consider as target variables $\mX_t$ only air temperature and irradiance (masked during night hours), while we use the remaining weather variables as exogenous $\mU_t$. This scenario differs from the one considered in previous experiments as the exogenous variables here vary across nodes rather than being the same temporal encodings.
\autoref{tab:engrad_exp} shows the experiment results. We can see that the results are consistent with~\autoref{tab:env} and both \gls{model} variants outperform all the baselines in every considered setting.

\section{Traffic Forecasting}
\label{sec:traffic_experiment}
In~\autoref{sec:scalability}, we pointed out at potential issues of using non-trainable pooling operators when the input graph is noisy or does not reflect the actual correlations and dependencies across the time series. In this experiment, we show this phenomenon in two popular traffic forecasting benchmarks, namely, \textbf{\gls{la}} and \textbf{\gls{bay}}. These datasets contain traffic speed records measured every $5$ minutes by $207$ and $325$ sensors located over highways in Los Angeles and the Bay Area, respectively. In both datasets, the adjacency matrix is directed and, as discussed in~\appref{sec:model_appendix}, it must be transformed to undirected to obtain the coarsened graphs when using the $k$-MIS pooling method.

\begin{table*}[t]
\caption{Forecasting error on traffic benchmarks with different block missing settings. \textbf{Bold} formatting is used to mark best result in each setting. $^\dagger$Models without spatial message passing.}
\vskip 0.15in
\label{tab:traffic}
\setlength{\tabcolsep}{5.5pt}
\setlength{\aboverulesep}{0pt}
\setlength{\belowrulesep}{0pt}
\renewcommand{\arraystretch}{1.2}
\begin{center}
\begin{small}
\begin{tabular}{l|cc|cc|cc|cc}
\cmidrule[.7pt]{2-9} 
\multicolumn{1}{c}{} & \multicolumn{4}{c|}{\bfseries \gls{la}} & \multicolumn{4}{c}{\bfseries \gls{bay}} \\
\cmidrule[.7pt]{2-9} 
\multicolumn{1}{c}{} & \multicolumn{2}{c|}{\bfseries \gls{block}} & \multicolumn{2}{c|}{\bfseries \gls{block_prop}} & \multicolumn{2}{c|}{\bfseries \gls{block}} & \multicolumn{2}{c}{\bfseries \gls{block_prop}} \\
\toprule
\multicolumn{1}{c|}{\textbf{Model}} & MAE & MSE & MAE & MSE & MAE & MSE & MAE & MSE \\
\midrule
\gls{grud}$^\dagger$ & 3.91{{\tiny$\pm$0.03}} & 62.03{{\tiny$\pm$1.06}} & 4.17{{\tiny$\pm$0.04}} & 70.11{{\tiny$\pm$1.09}} & 2.10{{\tiny$\pm$0.01}} & 24.19{{\tiny$\pm$0.17}} & 2.19{{\tiny$\pm$0.01}} & 26.08{{\tiny$\pm$0.29}} \\
\gls{grui}$^\dagger$ & 4.13{{\tiny$\pm$0.00}} & 69.47{{\tiny$\pm$0.24}} & 4.49{{\tiny$\pm$0.01}} & 80.43{{\tiny$\pm$0.53}} & 2.16{{\tiny$\pm$0.01}} & 25.54{{\tiny$\pm$0.05}} & 2.25{{\tiny$\pm$0.00}} & 27.89{{\tiny$\pm$0.01}} \\
\gls{grin} & 3.48{{\tiny$\pm$0.07}} & 49.59{{\tiny$\pm$1.51}} & 3.67{{\tiny$\pm$0.06}} & 55.14{{\tiny$\pm$0.78}} & 1.93{{\tiny$\pm$0.03}} & 18.33{{\tiny$\pm$0.61}} & 2.00{{\tiny$\pm$0.04}} & 19.60{{\tiny$\pm$0.51}} \\
\midrule
\gls{gru}$^\dagger$ & 4.01{{\tiny$\pm$0.01}} & 67.26{{\tiny$\pm$0.34}} & 4.31{{\tiny$\pm$0.01}} & 76.29{{\tiny$\pm$0.54}} & 2.08{{\tiny$\pm$0.00}} & 24.29{{\tiny$\pm$0.08}} & 2.17{{\tiny$\pm$0.00}} & 26.40{{\tiny$\pm$0.17}} \\
\gls{dcrnn} & 3.48{{\tiny$\pm$0.01}} & 47.21{{\tiny$\pm$0.48}} & 3.78{{\tiny$\pm$0.03}} & 56.49{{\tiny$\pm$1.39}} & 1.86{{\tiny$\pm$0.00}} & 17.66{{\tiny$\pm$0.07}} & 1.95{{\tiny$\pm$0.00}} & 19.39{{\tiny$\pm$0.14}} \\
\gls{agcrn} & 3.38{{\tiny$\pm$0.00}} & 47.80{{\tiny$\pm$0.74}} & 3.53{{\tiny$\pm$0.01}} & 52.89{{\tiny$\pm$0.77}} & 1.80{{\tiny$\pm$0.01}} & 16.56{{\tiny$\pm$0.33}} & 1.85{{\tiny$\pm$0.01}} & 17.41{{\tiny$\pm$0.15}} \\
\gls{gwnet} & 3.21{{\tiny$\pm$0.01}} & \textbf{42.17{{\tiny$\pm$0.47}}} & 3.43{{\tiny$\pm$0.03}} & 49.23{{\tiny$\pm$1.66}} & 1.71{{\tiny$\pm$0.01}} & 15.30{{\tiny$\pm$0.21}} & 1.78{{\tiny$\pm$0.00}} & 16.69{{\tiny$\pm$0.08}} \\
\gls{tasimp} & 3.28{{\tiny$\pm$0.01}} & 45.61{{\tiny$\pm$0.39}} & 3.45{{\tiny$\pm$0.01}} & 51.63{{\tiny$\pm$0.70}} & 1.74{{\tiny$\pm$0.00}} & 15.81{{\tiny$\pm$0.08}} & 1.81{{\tiny$\pm$0.01}} & 17.24{{\tiny$\pm$0.05}} \\
\gls{tasamp} & 3.24{{\tiny$\pm$0.01}} & 44.67{{\tiny$\pm$0.45}} & 3.45{{\tiny$\pm$0.01}} & 52.32{{\tiny$\pm$0.82}} & 1.74{{\tiny$\pm$0.00}} & 15.94{{\tiny$\pm$0.16}} & 1.81{{\tiny$\pm$0.00}} & 17.34{{\tiny$\pm$0.07}} \\
\gls{ttsimp} & \textbf{3.16{{\tiny$\pm$0.02}}} & 42.36{{\tiny$\pm$0.96}} & \textbf{3.36{{\tiny$\pm$0.02}}} & \textbf{49.01{{\tiny$\pm$0.58}}} & \textbf{1.70{{\tiny$\pm$0.00}}} & 14.94{{\tiny$\pm$0.17}} & 1.77{{\tiny$\pm$0.01}} & 16.34{{\tiny$\pm$0.15}} \\
\gls{ttsamp} & 3.19{{\tiny$\pm$0.02}} & 43.40{{\tiny$\pm$1.17}} & 3.40{{\tiny$\pm$0.02}} & 50.89{{\tiny$\pm$0.79}} & 1.74{{\tiny$\pm$0.00}} & 15.76{{\tiny$\pm$0.13}} & 1.80{{\tiny$\pm$0.01}} & 17.18{{\tiny$\pm$0.21}} \\
\cmidrule[.8pt]{1-9}
\textbf{\gls{model}-IMP} & 3.23{{\tiny$\pm$0.01}} & 44.11{{\tiny$\pm$0.55}} & 3.40{{\tiny$\pm$0.01}} & 50.18{{\tiny$\pm$0.39}} & \textbf{1.70{{\tiny$\pm$0.01}}} & 15.01{{\tiny$\pm$0.30}} & \textbf{1.76{{\tiny$\pm$0.01}}} & 16.17{{\tiny$\pm$0.25}} \\
\textbf{\gls{model}-AMP} & 3.22{{\tiny$\pm$0.01}} & 43.41{{\tiny$\pm$0.36}} & 3.39{{\tiny$\pm$0.02}} & 49.62{{\tiny$\pm$0.62}} & \textbf{1.70{{\tiny$\pm$0.01}}} & \textbf{14.91{{\tiny$\pm$0.21}}} & 1.77{{\tiny$\pm$0.00}} & \textbf{16.08{{\tiny$\pm$0.12}}} \\
\bottomrule
\end{tabular}
\end{small}
\end{center}
\vskip -0.1in
\end{table*}

\begin{figure}[h]
    \centering
    \hfill
    \includegraphics[width=.48\textwidth]{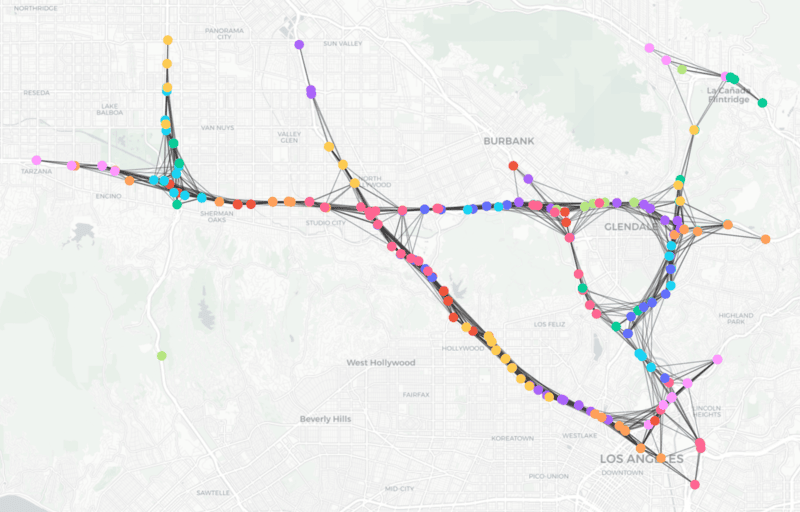}
    \hfill
    \includegraphics[width=.48\textwidth]{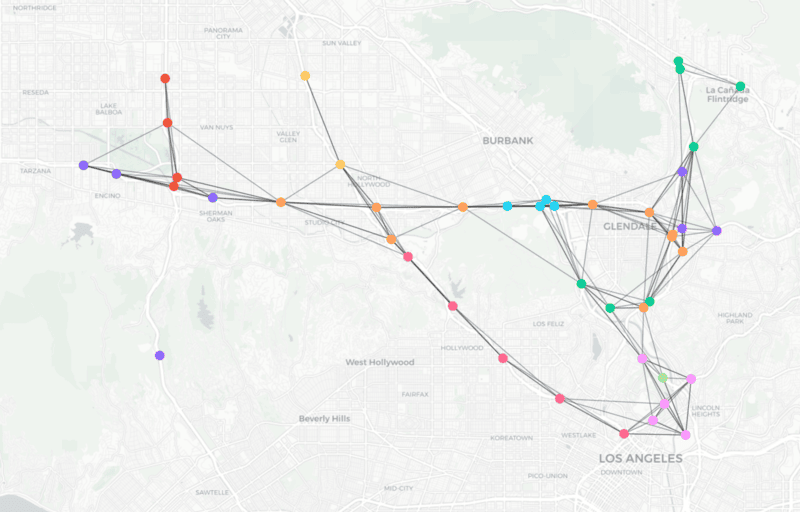}
    \hfill
    ~\\[.5cm]~
    \hfill
    \includegraphics[width=.48\textwidth]{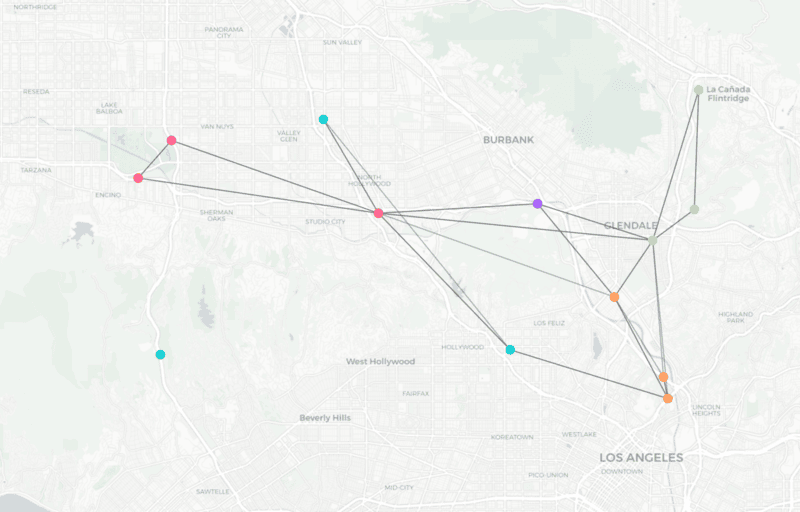}
    \hfill
    \includegraphics[width=.48\textwidth]{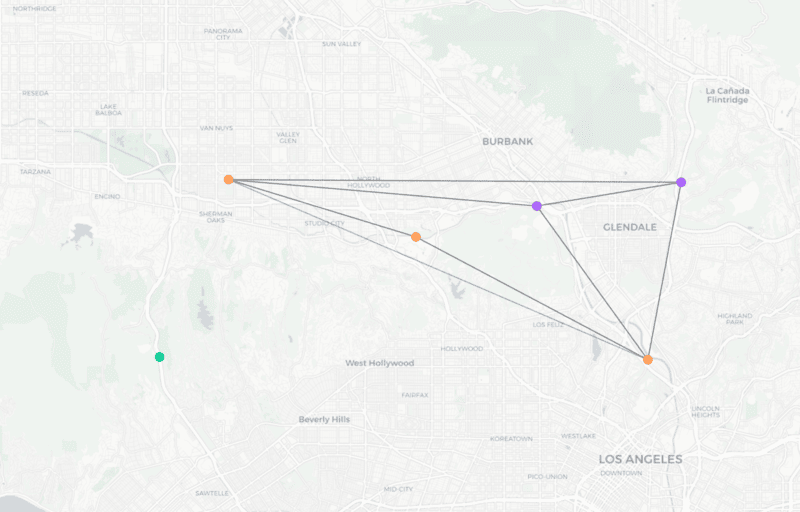}
    \hfill~
    \caption{Coarsened graphs in \gls{la} obtained by recursive application of $k$-MIS pooling.}
    \label{fig:la_pooled}
\end{figure}

As it is evident from~\autoref{fig:la_pooled}, the graph topology does not reflect the underlying structure of the road network. Distant nodes are assigned by the pooling operator to the same supernode, causing significant discrepancies between the topology of the coarsened graphs and the underlying data. Indeed, points at a significant driving distance end up being connected in the graph. Additionally, some nodes are disconnected from the rest of the graph ($5$ in \gls{la} and $12$ in \gls{bay}). These issues affect the computation of the coarsened graphs and, consequently, the performance of \gls{model}, which relies on spatially aggregated representations to capture long-range spatial dynamics.
\autoref{tab:traffic} shows the performance of the models. 
While outperforming the baselines in some settings, in some other cases flatter models (like \gls{ttsimp}) are more effective.

\end{document}